\documentclass[10pt,twocolumn,letterpaper]{article}

\usepackage{cvpr}              %

\usepackage{graphicx}
\usepackage{amsmath}
\usepackage{amssymb}
\usepackage{booktabs}
\usepackage{cuted}
\usepackage{gensymb}
\usepackage{multirow}
\usepackage[table]{xcolor} 
\definecolor{best1}{RGB}{222,242,212}
\definecolor{best2}{RGB}{255,250,212}
\usepackage[pagebackref,breaklinks,colorlinks]{hyperref}
\usepackage[accsupp]{axessibility}  %

\usepackage[capitalize]{cleveref}
\crefname{section}{Sec.}{Secs.}
\Crefname{section}{Section}{Sections}
\Crefname{table}{Table}{Tables}
\crefname{table}{Tab.}{Tabs.}

\def\cvprPaperID{3019} %
\def\confName{CVPR}
\def\confYear{2023}

\begin{document}

\title{Neural Residual Radiance Fields for Streamably Free-Viewpoint Videos}

\author{Liao Wang$^{1,3}$
\and
Qiang Hu$^{1}$
\and
Qihan He$^{1,4}$
\and
Ziyu Wang$^1$
\and
Jingyi Yu$^{1}$
\and
Tinne Tuytelaars$^{2}$
\qquad \qquad Lan Xu$^{1}$$^{\dagger}$  \qquad \qquad
Minye Wu$^{2}$$^{\dagger}$   \\
{$^{1}$ ShanghaiTech University \qquad $^{2}$ KU Leuven \qquad $^{3}$ NeuDim \qquad $^{4}$ DGene} \\
}

\maketitle

\begin{strip}\centering
\vspace{-17mm}
	\includegraphics[width=\linewidth]{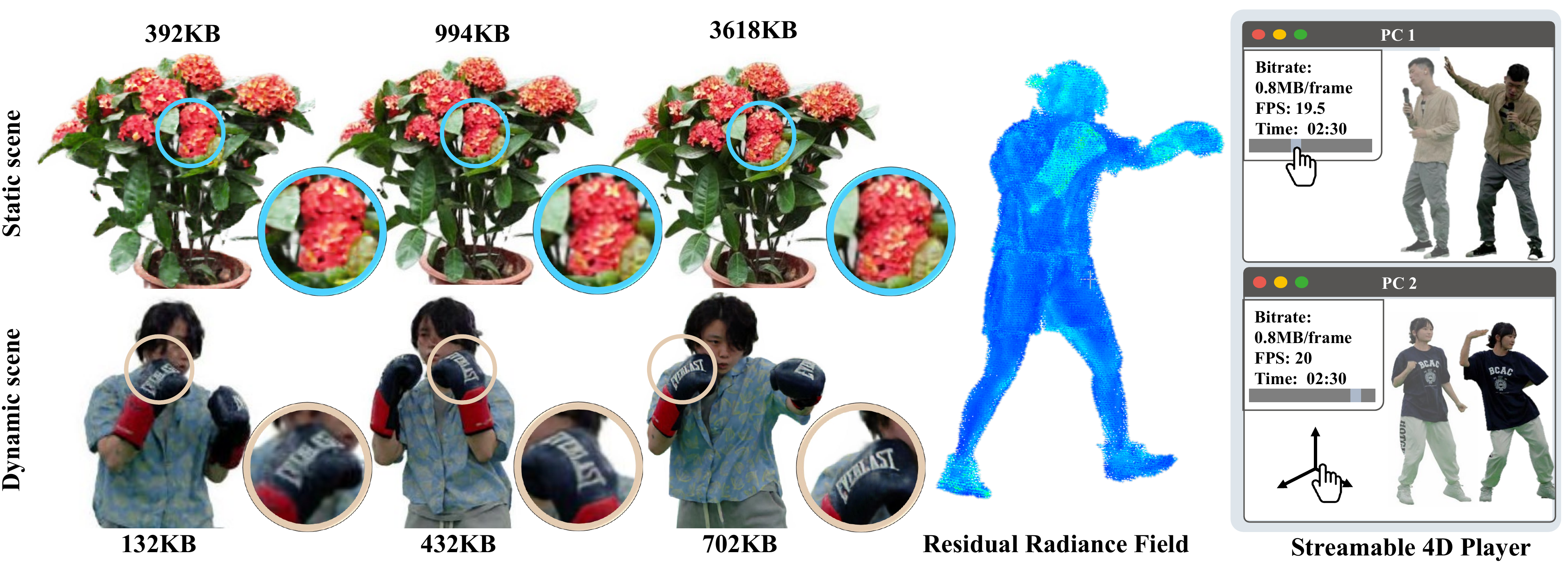}
	\vspace{-0.5cm}
	\captionof{figure}{Our proposed ReRF utilizes a residual radiance field and a global MLP to enable highly compressible and streamable radiance field modeling. Our ReRF-based codec scheme and streaming player gives users a rich interactive experience.} %
	\label{fig:teaser}
\end{strip}

{\let\thefootnote\relax\footnote{\
		${ }^{\dagger}$ The corresponding authors are Minye Wu
		(minye.wu@kuleuven.be) and Lan Xu (xulan1@shanghaitech.edu.cn).
}}\par

\begin{abstract}
     The success of the Neural Radiance Fields (NeRFs) for modeling and free-view rendering static objects has inspired numerous attempts on dynamic scenes. Current techniques that utilize neural rendering for facilitating free-view videos (FVVs) are restricted to either offline rendering or are capable of processing only brief sequences with minimal motion.
    In this paper, we present a novel technique, Residual Radiance Field or ReRF, as a highly compact neural representation to achieve real-time FVV rendering on long-duration dynamic scenes. ReRF explicitly models the residual information between adjacent timestamps in the spatial-temporal feature space, with a global coordinate-based tiny MLP as the feature decoder.
    Specifically, ReRF employs a compact motion grid along with a residual feature grid to exploit inter-frame feature similarities. We show such a strategy can handle large motions without sacrificing quality. 
    We further present a sequential training scheme to  maintain the smoothness and the sparsity of the motion/residual grids.
    Based on ReRF, we design a special FVV codec that achieves three orders of magnitudes compression rate and provides a companion ReRF player to support online streaming of long-duration FVVs of dynamic scenes.    
    Extensive experiments demonstrate the effectiveness of ReRF for compactly representing dynamic radiance fields, enabling an unprecedented free-viewpoint viewing experience in speed and quality.

\end{abstract}

\section{Introduction} \label{sec:intro}

Photo-realistic free-viewpoint videos (FVVs) of dynamic scenes, in particular, human performances, reduce the gap between the performer and the viewer. But the goal of producing and viewing FVVs as simple as clicking and viewing regular 2D videos on streaming platforms remains far-reaching. The challenges range from data processing and compression to streaming and rendering.

Geometry-based solutions reconstruct dynamic 3D meshes or points ~\cite{collet2015high,motion2fusion}, whereas image-based ones interpolate novel views on densely transmitted footages~\cite{zitnick2004high,broxton2020immersive}. Both techniques rely on high-quality reconstructions that are often vulnerable to occlusions and textureless regions.
Recent neural advances~\cite{mildenhall2020nerf,NR_survey} bring an alternative route that bypasses explicit geometric reconstruction.
The seminal work of the Neural Radiance Field (NeRF)~\cite{mildenhall2020nerf} compactly represents a static scene in a coordinate-based multi-layer perceptron (MLP) to conduct volume rendering at photo-realism. The MLP can be viewed as an implicit feature decoder from a spatially continuous feature space to the radiance output with RGB and density. However, using even a moderately deep MLP can be too expensive for real-time rendering. Various extensions have hence focused on ``sculpting'' the feature space using smart representations to strike an intricate balance between computational speed and accuracy. Latest examples include explicit feature volumes~\cite{fridovich2022plenoxels,yu2021plenoctrees,sun2021direct}, multi-scale hashing~\cite{muller2022instant}, codebook~\cite{takikawa2022variable}, tri-planes~\cite{EG3D}, tensors~\cite{chen2022tensorf,tang2022compressible}, etc.

Although effective, by far nearly all methods are tailored to handle static scenes. In contrast, streaming dynamic radiance fields require using a global coordinate-based MLP to decode features from a spatial-temporally continuous feature space into radiance outputs.
A na\"{i}ve per-frame solution would be to apply static methods ~\cite{tang2022compressible,muller2022instant} on a series of independent spatial feature spaces. Such schemes discard important temporal coherency, yielding low quality and inefficiency for long sequences. Recent methods attempt to maintain a canonical feature space to reproduce features in each live frame by temporally warping them back into the canonical space. Various schemes to compensate for temporal motions have been proposed by employing implicit matching~\cite{tretschk2020non,park2020deformable,park2021hypernerf,liu2022devrf,fang2022fast} or data-driven priors such as depth~\cite{xian2020space}, Fourier features~\cite{wang2022fourier}, optical flow~\cite{du2021nerflow,li2020neural}, or skeletal/facial motion priors~\cite{peng2021neural,zhao2022humannerf,weng2022humannerf,hong2022headnerf}. However, heavy reliance on the global canonical space makes them fragile to large motions or topology changes.  The training overhead also significantly increases according to the sequence length.
Recent work~\cite{li2022streaming} sets out to explore feature redundancy between adjacent frames but it falls short of maintaining a coherent spatial-temporal feature space.

In this paper, we present a novel neural modeling technique that we call the Residual Radiance Field or ReRF as a highly compact representation of dynamic scenes, enabling high-quality FVV streaming and rendering (Fig.~\ref{fig:teaser}). ReRF explicitly models the residual of the radiance field between adjacent timestamps in the spatial-temporal feature space. 
Specifically, we employ a global tiny MLP to approximate radiance output of the dynamic scene in a sequential manner. To maintain high efficiency in training and inference, ReRF models the feature space using an explicit grid representation analogous to ~\cite{sun2021direct}. 
However, ReRF only performs the training on the first key frame to obtain an MLP decoder for the whole sequence and at the same time it uses the resulting grid volume as the initial feature volume.
For each subsequent frame, ReRF uses a compact motion grid and a residual feature grid: the low-resolution motion grid represents the position offset from the current frame to the previous whereas a sparse residual grid is used to compensate for errors and newly observed regions. A major benefit of such a design is that ReRF fully exploits feature similarities between adjacent frames where the complete feature grid of the current frame can be simply obtained from the two while avoiding the use of a global canonical space. In addition, both motion and residual grids are amenable for compression, especially for long-duration dynamic scenes.

\begin{figure}[t]
\vspace {-3mm}
\begin{center}
    \includegraphics[width=\linewidth]{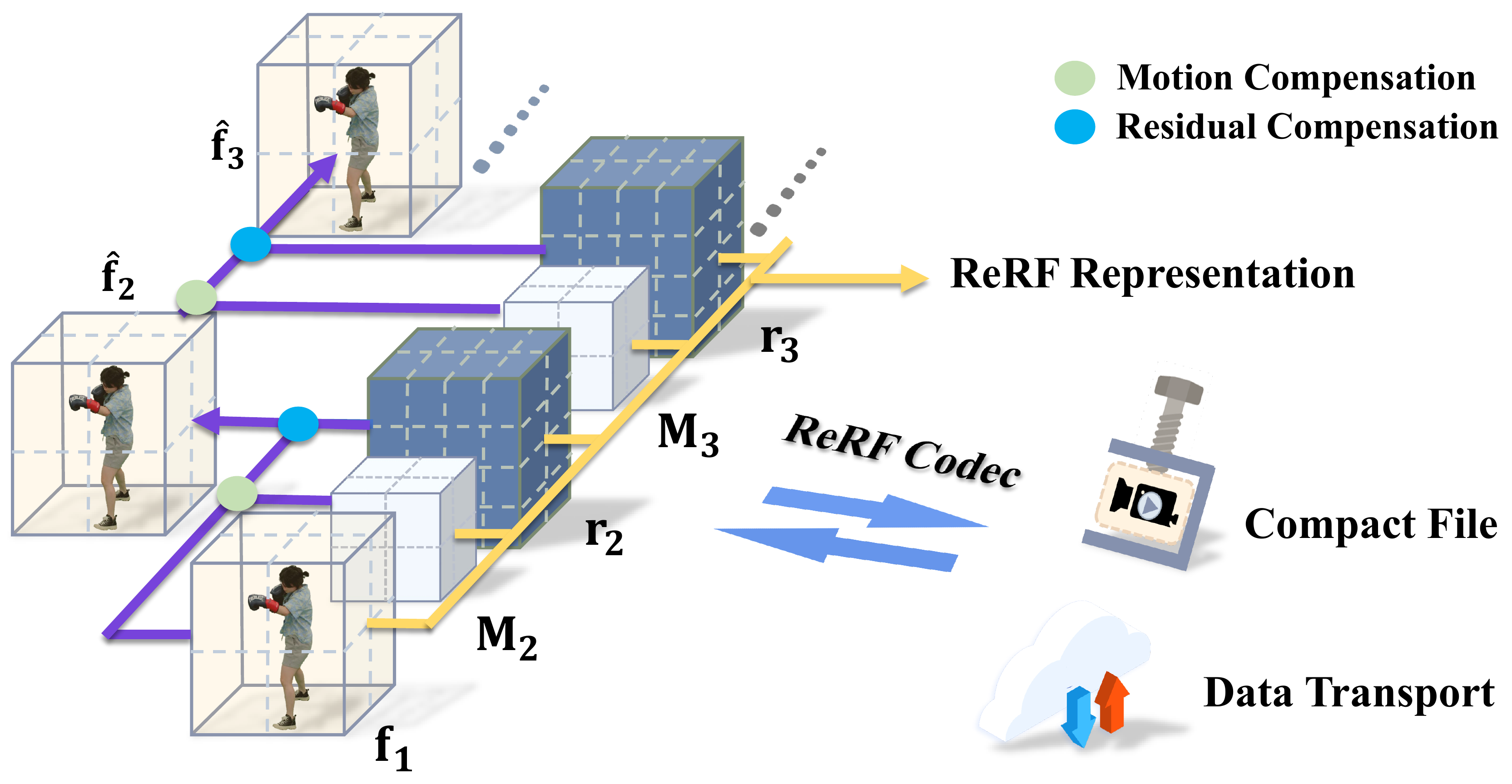}
\end{center}
\vspace {-5mm}
\caption{Overview of our method. We first use our sequential training scheme (Sec.~\ref{sec:method}) to generate compact ReRF representation with motion grid $\mathbf{M}_i$ and $\mathbf{r}_i$ for each frame $i$. Next, our ReRF-based codec scheme and player (Sec.~\ref{sec:4}) will compress it to enable fast data transport and online playing.}
\label{fig:pipeline}
\vspace{-5mm}
\end{figure}

We present a two-stage scheme to efficiently obtain the ReRF from RGB videos via sequential training. In particular, we introduce a novel motion pooling strategy to maintain the smoothness and compactness of the inter-frame motion grid along with sparsity regularizers to improve the compactness of ReRF.
To make ReRF practical for users, we further design a ReRF-based codec that follows the traditional keyframe-based strategy, achieving three orders of magnitudes compression rate compared to per-frame-based neural representations~\cite{sun2021direct}. Finally, we demonstrate a companion ReRF player suitable for conducting online streaming of long-duration FVVs of dynamic scenes. With ReRF, a user, for the first time, can pause, play, fast forward/backward, and seek on dynamic radiance fields as if viewing 2D videos, resulting in an unprecedented high-quality free-viewpoint viewing experience (see Fig.~\ref{fig:pipeline}).

To summarize, our contributions include:
\begin{itemize}
	\item We introduce Residual Radiance Field (ReRF), a novel neural representation, to support streamable free-viewpoint viewing of dynamic radiance fields.
	
	\item We present tailored motion and residual grids to support sequential training and at the same time eliminate the need for using a global canonical space notorious for large motions. We further introduce a number of training strategies to achieve a high compression rate while maintaining high rendering quality.

	\item We develop a ReRF-based codec and a companion FVV player to stream dynamic radiance fields of long sequences, with broad control functions.

\end{itemize}

\section{Related work}
\label{sec:Relatedwork}

\textbf{Novel View Synthesis for Static Scenes.}
Novel view synthesis, the problem of synthesizing new viewpoints given a set of 2D images,  has recently attracted considerable attention. 
Light field representations~\cite{levoy1996light, gortler1996lumigraph, chai2000plenoptic,feng2021signet, attal2022learning} formulates the problem by two-plane parametrization. Early methods~\cite{levoy1996light, gortler1996lumigraph, chai2000plenoptic} generate rays of a novel viewpoint via interpolation, which can achieve real-time rendering but require caching all rays. 
Recent works~\cite{feng2021signet, attal2022learning} use neural networks for compact storage.
Mesh-based representations~\cite{wood2000surface,waechter2014let,chen2018deep} allow for efficient storage and can record the view-dependent texture~\cite{wood2000surface,chen2018deep}. However, optimizing a mesh to fit a scene with complex topology is still a challenge.
Multi-plane images~\cite{szeliski1998stereo,penner2017soft,choi2019extreme,flynn2019deepview, wizadwongsa2021nex} have shown the ability to handle complex scenes because of their topology-free nature.
More recently, the breakthrough approach NeRF~\cite{mildenhall2020nerf} greatly improves the realism of rendering and inspires numerous follow-up works including multi-scale~\cite{barron2021mip,barron2022mip}, relighting~\cite{boss2021nerd, srinivasan2021nerv, zhang2021nerfactor}, editing~\cite{yuan2022nerf,yang2021objectnerf}, 3D-aware generation~\cite{gu2021stylenerf,EG3D,deng2022gram,wang2022generative,poole2022dreamfusion}, etc.
However, ~\cite{mildenhall2020nerf} assumes a static scene and cannot handle scene variations over time.

\textbf{Novel View Synthesis for Dynamic Scenes.}
Dynamic scenes are more complex because of illumination variations and object movements.
One way is to reconstruct the dynamic scene and render the geometry from novel views. RGB~\cite{kim2010dynamic, collet2015high,ranftl2016dense,lv2018learning,li2019learning,luo2020consistent, zhao2022human} or RGB-D~\cite{motion2fusion,TotalCapture,UnstructureLan,newcombe2015dynamicfusion,FlyFusion,jiang2022neuralfusion} solutions have been widely explored.
Other methods~\cite{lombardi2019neural, Wu_2020_CVPR, bemana2020x} model the dynamic scene by neural networks for view synthesis. \cite{bemana2020x} use a neural network to regress each image from all others to achieve view, time, or light interpolation. \cite{lombardi2019neural} use an encoder-decoder network to transfer the 2D images into 3D volume, and leverages volumetric rendering for end-to-end training. \cite{Wu_2020_CVPR} combines the points feature with multi-view images for dynamic human rendering. Using motion-advected feature vectors \cite{Holynski_2021_CVPR} for still image animation is also an interesting direction.

More recently, \cite{park2020deformable,pumarola2021d,li2020neural,xian2020space,tretschk2020non,ost2020neural,st-nerf,Gao-ICCV-DynNeRF,du2021nerflow,park2021hypernerf, wang2021dctnerf,fang2022fast,luo2022artemis,li2022dynerf, wang2022fourier, liu2022devrf,wang2021ibutter,li2020neural} extend NeRF~\cite{mildenhall2020nerf} into the dynamic settings. 
Some~\cite{xian2020space,Gao-ICCV-DynNeRF,du2021nerflow} directly condition the neural radiance field on time to handle spatial changes. 
Others~\cite{tretschk2020non,li2020neural,park2020deformable,pumarola2021d,st-nerf,zhao2022humannerf,li2022dynerf} learn spatial offsets from the current scene to a learned canonical radiance field at each timestamp. 
~\cite{park2021hypernerf} conditions NeRF on additional higher-dimensional coordinates to tackle the discontinuous topological changes beyond the continuous deformation field. 
~\cite{wang2021dctnerf} handles scene dynamic change by modeling the trajectory of each point in the scene. 
~\cite{liu2022devrf} uses explicit voxels to model both the canonical space and deformation field for dynamic scenes. ~\cite{wang2022fourier} models the time-varying density and color by Fourier coefficients to extend the octree-based radiance field~\cite{yu2021plenoctrees} to dynamic scenes. Compared to \cite{wang2022fourier}, our method uses three orders of magnitude smaller storage and enables long sequences with large motions.

\textbf{NeRF Acceleration and Compression.}
NeRF~\cite{mildenhall2020nerf} shows extraordinary results in free-view rendering, but its training and rendering speed are slow. Recent approaches reduce the complex MLP computation by decomposing NeRF into explicit 3D feature encoding with a shallow MLP decoder. Methods have been explored involving voxel grids~\cite{nsvf,hedman2021baking,sun2021direct,li2022streaming},
octrees~\cite{yu2021plenoctrees,fridovich2022plenoxels, wang2022fourier}, tri-planes~\cite{EG3D}, multi-scale hashing~\cite{muller2022instant}, codebook~\cite{takikawa2022variable}, tensor decomposition~\cite{chen2022tensorf,tang2022compressible,song2022nerfplayer}, and textured polygons~\cite{chen2022mobilenerf}.

Using explicit encoding greatly reduces training and inference time, but the additional storage consumption associated with these 3D structures is a concern.
Some methods achieve high compression ratios through CP-decomposition~\cite{chen2022tensorf}, rank reduction~\cite{tang2022compressible} or vector quantization~\cite{takikawa2022variable} but are limited to static scenes.
Recent dynamic approaches~\cite{li2022streaming} employ narrow band tuning on sparse voxel grids for video sequences, which is efficient to train but still has a size of MB per frame. ~\cite{song2022nerfplayer} decomposes the 4D space into static, deforming, and new areas for efficient dynamic scene training and  rendering, but is limited by the length of the video sequence. 
In contrast, we embrace residual radiance field and ReRF-based codec scheme, which enables high compression and streaming for long sequences with large motion.

\section{Neural Residual Radiance Field} \label{sec:method}

\begin{figure*}[t]
\vspace{-1.0cm}
	\begin{center}
		\includegraphics[width=1.0\linewidth]{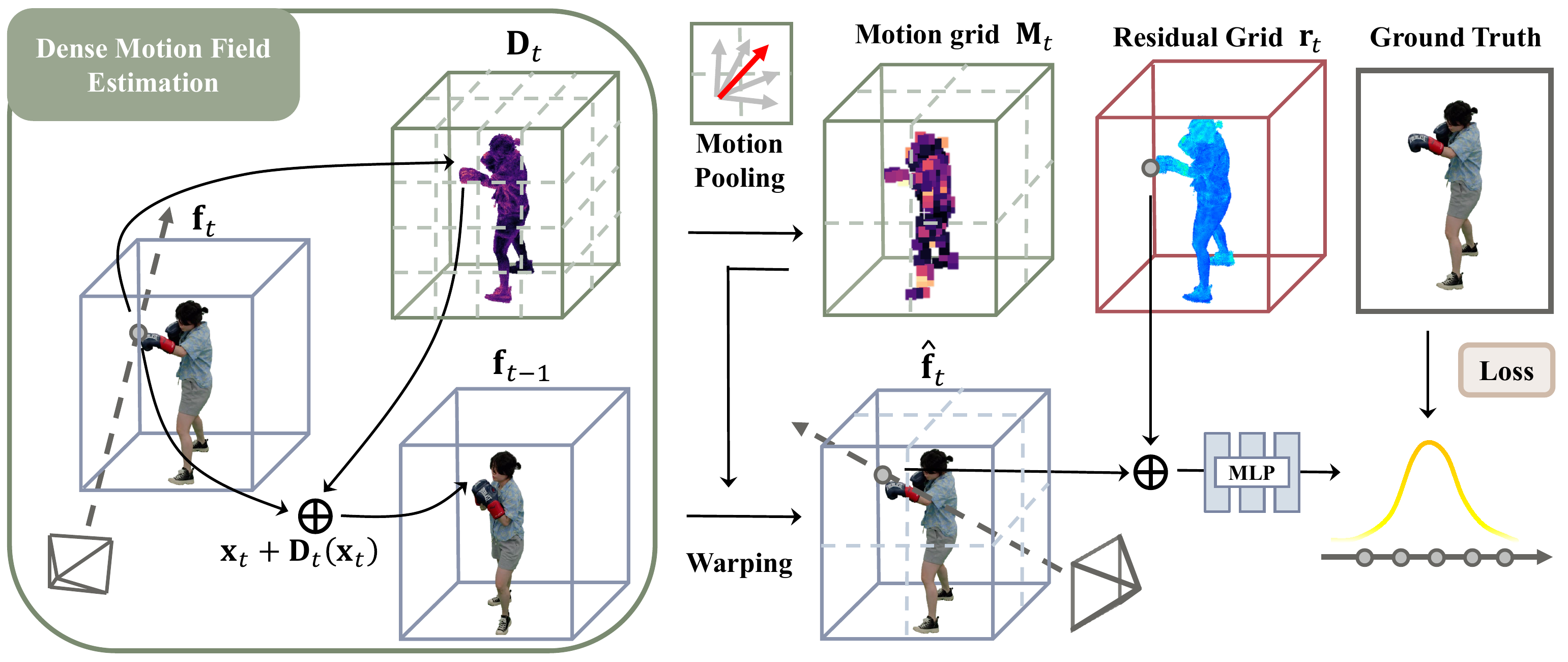}
	\end{center}
	\vspace{-0.5cm}
	\caption{ Illustration of our Neural Residual Radiance Field (ReRF). First, we estimate a dense motion field $\mathbf{D}_t$. Next, we generate a compact motion grid $\mathbf{M}_t$ through motion pooling. Finally, we warp $\mathbf{f}_{t-1}$ to a base grid $\mathbf{\hat f}_{t}$ and learn our residual grid $\mathbf{r}_t$ to increase feature sparsity and promote compression.}
	\label{fig:method}
	\vspace{-5mm}
\end{figure*}

In this section, we introduce the details about the proposed ReRF representation for dynamic scenes (Sec.~\ref{sec:Revoxel}), followed by a companion training scheme to generate ReRF from RGB video inputs (Sec.~\ref{sec:training}).

\subsection{Motion-aware Residual Fields}  \label{sec:Revoxel}
Recall that the radiance with color and density $(\mathbf{c},\sigma)$ in NeRF is formulated as $\mathbf{c},\sigma = \Psi(\mathbf{x},\mathbf{d})$, using MLPs as decoder given the 3D position $\mathbf{x}$ and viewing direction $\mathbf{d}$. 
Then, volume rendering is adopted for photo-realistic novel view synthesis based on the radiance fields.
To maintain high efficiency in training and inference, in ReRF,  we use an explicit grid representation similar to previous work~\cite{sun2021direct}.
Specifically, with an explicit density grid $\mathbf{V}_{\sigma}$ and a color feature grid $\mathbf{V}_{c}$, the radiance field of a static scene is:
\begin{equation}
\begin{split}
\sigma &= interp(\mathbf{x},\mathbf{V}_{\sigma}) \\
\mathbf{c} &= \Phi(interp(\mathbf{x},\mathbf{V}_{c}),\mathbf{d}), 
\end{split}
\label{eq:density}
\end{equation}
where $interp(\cdot)$ denotes the trilinear interpolation function on the grids, and $\Phi$ is a relatively shallow MLP for acceleration. 
For simplification, we can union $\mathbf{V}_{\sigma}$ and $\mathbf{V}_{c}$ into a common feature grid $\mathbf{f}$ by appending an additional channel to $\mathbf{V}_{c}$. 
To that end, the explicit grid representation for a static radiance field consists of a feature grid $\mathbf{f}$ and a tiny MLP $\Phi$ as the implicit feature decoder.

To further represent a dynamic radiance field, we adopt a coordinated-based tiny MLP $\Phi$ as the global feature decoder for the spatial-temporal feature space. A na\"{i}ve solution would be to utilize per-frame feature grids $\{\mathbf{f}_t\}_{t=1}^N$ for the dynamic scene with N frames, yet discarding important temporal coherency. Recent work DeVRF~\cite{liu2022devrf} maintains a canonical feature grid $\mathbf{f}_1$ with dense motion fields $\{\mathbf{D}_t\}_{t=1}^N$ to reproduce features in each live frame, but it's fragile to large motions or topology changes due to the reliance on a canonical space.
In stark contrast, we propose to explicitly exploit the feature similarities between adjacent timestamps in the spatial-temporal feature space. Here, we introduce a compact motion grid $\mathbf{M}_t$ and a residual feature grid $\mathbf{r}_t$ for the current frame $t$. The low-resolution motion grid $\mathbf{M}_t$ denotes the voxel offset to indicate the corresponding voxel index in the previous frame for a voxel in the current frame. The residual grid $\mathbf{r}_t$ denotes the sparse compensation for both the adjacent warping error and the newly observed regions in the current frame. Besides, for the first frame, we adopt a complete explicit feature grid representation $\mathbf{f}_1$ with the companion global MLP $\Phi$.
Finally, our ReRF sequentially represents a dynamic radiance field with N frames as $\Phi$, $\mathbf{f}_1$, and $\{\mathbf{M}_t,\mathbf{r}_t\}_{t=1}^N$, as illustrated in Fig.~\ref{fig:pipeline}.    

Note that our ReRF enables highly efficient sequential feature modeling. Given the previous $\mathbf{f}_{t-1}$, current feature grid $\mathbf{f}_{t}$ can be simply obtained from $\mathbf{M}_t$ and $\mathbf{r}_t$ while avoiding the use of global canonical space. Specifically, we first apply $\mathbf{M}_t$ to $\mathbf{f}_{t-1}$ to extract the inter-frame redundancy and obtain a base feature grid $\mathbf{\hat f}_{t}$ for the current frame. Let $\mathbf{p}$ denote the index of our explicit grids. Then, the per-voxel base feature grid is formulated as: 
\begin{equation}
\begin{split}
 \mathbf{\hat f}_{t}(\mathbf{p}) = \mathbf{f}_{t-1}(\mathbf{p} + \mathbf{M}_t(\mathbf{p})),
\end{split}
\vspace{-3mm}
\label{eq:basev}
\end{equation}
which turns to exploiting the inter-frame feature similarities as much as possible. 
We then recover the entire feature grid by adding the residual compensation: $\mathbf{f}_t = \mathbf{\hat f}_{t} +\mathbf{r}_t$, enabling the reconstruction of the current radiance field by applying the global MLP $\Phi$ on $\mathbf{f}_t$ according to Eqn~\ref{eq:density}.
Compared to the explicit feature grids $\{\mathbf{f}_t\}$, our motion-aware residual representation $\{\mathbf{M}_t,\mathbf{r}_t\}$ is compact and compression-friendly, which naturally models feature changes in the coherent spatial-temporal feature space.

\subsection{Sequential Residual Field Generation}  \label{sec:training}
Here, we introduce a two-stage and sequential training scheme to obtain a ReRF representation including $\Phi$, $\mathbf{f}_1$, and $\{\mathbf{M}_t,\mathbf{r}_t\}_{t=1}^N$ from long-duration RGB video inputs, which naturally enforces the compactness of both residual and motion grids to enable the fascinating streamable applications in Sec.~\ref{sec:4}. 
At the very beginning, we utilize the off-the-shelf approach~\cite{sun2021direct} to obtain the complete explicit feature grid $\mathbf{f}_1$ for the first frame, companion with the global MLP $\Phi$ as feature decoder. Then, sequentially given the feature grid $\mathbf{f}_{t-1}$ of the previous frame and the input images for the current frame, we compactly generate the motion grid $\mathbf{M}_t$ and residual grid $\mathbf{r}_t$ in the following two stages.

\textbf{Motion Grid Estimation.} 
We first follow DeVRF~\cite{liu2022devrf} to a dense motion field $\mathbf{D}_t$ yet only from the current frame to the previous one by treating the previous frame as the canonical space. To maintain a smooth and compact motion grid $\mathbf{M}_t$, we further introduce a motion pooling strategy. 
Motion vectors in a voxel $\mathbf{p}_{t}$ may point to different voxels $\mathbf{p}_{t-1}$ in the previous frame. 
Thus, analogous  to the standard average pooling operation, we select the voxel $\mathbf{\bar p}_{t-1}$ that the mean vector points to as the voxel motion $\mathbf{M}_t(\mathbf{p}_{t}) = \mathbf{\bar p}_{t-1}$. 
Specifically, we first split the $\mathbf{D}_t$ into cubes, where each cube contains continuous $8\times 8 \times 8$ voxels.
Then, for each cube we apply an average pooling on the $\mathbf{D}_t$ at the kernel of $8\times 8 \times 8$, to enforce that each cube shares the same motion vector. After that, we downsample it to generate a low-resolution motion grid $\mathbf{M}_t$.
Note that our compact motion grid $\mathbf{M}_t$ is compression-friendly since its size is 512 times smaller than the original dense one.
In this way, some feature cubes from the former frame can be tracked through the motion field, so that the entropy of the residual voxels can be further decreased.
To that end, we generate a low-resolution $\mathbf{M}_t$ that compactly represents the smooth motions across frames.

\begin{figure}[t]
	\vspace{-0.7cm}
	\begin{center}
		\includegraphics[width=1.0\linewidth]{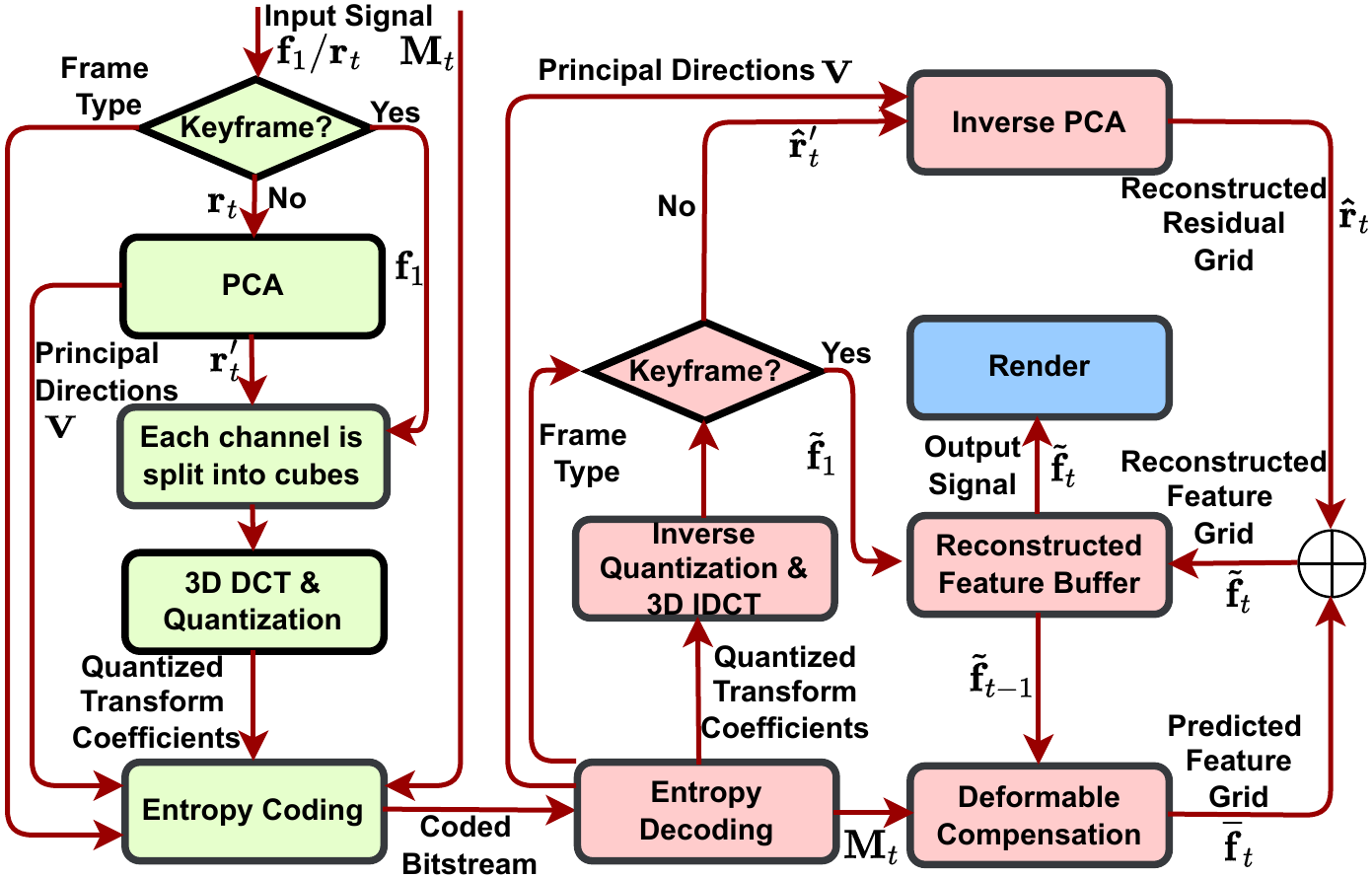}
	\end{center}
	\vspace{-0.5cm}
	\caption{Overview of our proposed ReRF-based codec and player (the modeling elements of the encoder and decoder are shaded in light green and pink, respectively). The encoder compresses the input signal to produce a bitstream by using PCA, 3D-DCT, quantization, and entropy coding. The decoder receives the compressed bitstream, decodes each of the syntax elements, and reverses the coding process. Additionally, given the decoded motion field $\mathbf{M}_t$ and the previously reconstructed feature grid $\mathbf{\widetilde{f}}_{t-1}$, we can obtain the predicted feature grid $\mathbf{\overline{f}}_t$ by deformation. }
	\label{fig:encoder}
	\vspace{-5mm}
\end{figure}

\textbf{Residual Grid Optimization.} 
With the aid of the compact motion grid $\mathbf{M}_t$, we warp previous feature grid $\mathbf{f}_{t-1}$ into the current base grid $\mathbf{\hat f}_{t}$, which coarsely compensates the feature differences caused by inter-frame motion. 
During optimizing the residual grid, we fix $\mathbf{\hat f}_{t}$ and $\Phi$ and back-propagate the gradients to the residual grid $\mathbf{r}_t$ to only update $\mathbf{r}_t$.
Apart from the photometric loss, we also regularize $\mathbf{r}_t$ by using an L1 loss to enhance its sparsity to improve compactness.
Such sparse formulation also enforces that $\mathbf{r}_t$ only compensates the sparse information for inter-frame residue or the newly observed regions. 
The total loss function $\mathcal{L}_{total}$ for learning $\mathbf{f}_t$ is formulated as:
\begin{equation}
\begin{split}
\mathcal{L}_{total} = \sum_{\mathbf{l} \in \mathbb{L}}\|\mathbf{c}(\mathbf{l})-\mathbf{\hat c}(\mathbf{l})\|^2 +  \lambda\|\mathbf{r}_t\|_1
\end{split}
\label{eq:loss}
\end{equation}
where $\mathbb{L}$ is the set of training pixel rays; $\mathbf{c}(\mathbf{l})$ and $\mathbf{\hat c}(\mathbf{l})$ are the ground truth color and predicted color of a ray $\mathbf{l}$ respectively; $\lambda=0.01$ is the weight of the regularization term.

Once obtained $\mathbf{M}_t, \mathbf{r}_t$, we can recover the explicit feature grid $\mathbf{f}_t$ of the current frame as illustrated in Sec.~\ref{sec:Revoxel}, and also enables the successive training of next frame. Note that the design and generation mechanism of $\mathbf{M}_t$ and $\mathbf{r}_t$ makes them compression-friendly due to their compact representation and sparse property, enabling following ReRF codec and streaming 
Please refer to our supplementary material for more training details of ReRF.

\section{ReRF Codec and Streamble Application}\label{sec:4}

\subsection{Feature-level Residual Compression.}
\label{sec:4.1}
Both motion and residual grids are amenable for compression, especially for long-duration dynamic scenes. To make ReRF practical for users, we further propose a ReRF-based codec and a companion FVV player for online streaming of long-duration dynamic scenes,  as shown in Fig. \ref{fig:encoder}. %
We first divide the feature grid sequence into several continuous groups of feature grids (GOF), which is a collection of successive grids as shown in Fig \ref{GOF}.
GOFs are comprised of an I-feature grid (keyframe) and a P-feature grid. Each GOF begins with an I-feature grid which is coded independently of all other feature grids. %
The p-feature grid contains a deformable compensated residual grid relative to the previous feature grid.
Let $\{\mathbf{f}_1,\mathbf{r}_2,\cdots,\mathbf{r}_{t-1},\mathbf{r}_t,\cdots\}$ denote a GOF, 
where $\mathbf{f}_1$ is the feature grid and $\mathbf{r}_t$ is the residual grid.

\begin{figure}[t]
\vspace{-0.7cm}
	\begin{center}
		\includegraphics[width=1.0\linewidth]{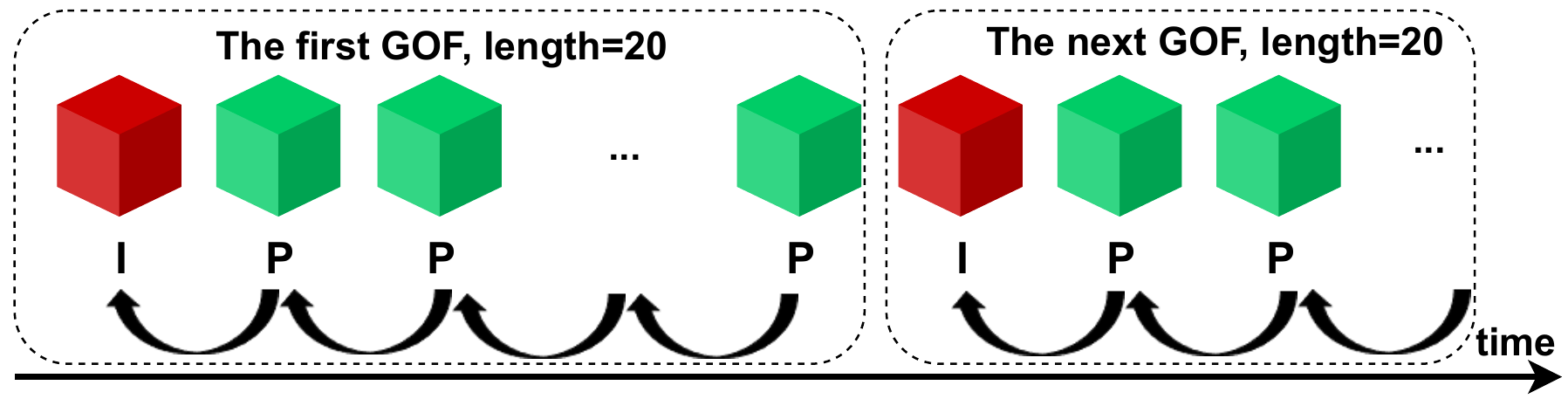}
	\end{center}
	\vspace{-0.6cm}
	\caption{ GOF structure. }
	\label{GOF}
	\vspace{-6mm}
\end{figure}

We first reshape $\mathbf{f}_1$ and $\mathbf{r}_t$ into $\mathbf{f}_1(m,n)$ and $\mathbf{r}_t(m,n)$, a $m\times n$ feature matrix, where $m$ and $n$ are the number of non-empty feature voxels and feature channels, respectively. %
Then, we perform linear Principal Component Analysis (PCA) \cite{pca} on $\mathbf{r}_t(m,n)$ to get principal directions $\mathbf{V}$. 
Finally, we project the $\mathbf{r}_t$ to principal directions by $\mathbf{r}_t^{\prime}=\mathbf{r}_t\cdot \mathbf{V}$. Each channel of grid $\mathbf{f}_1$ and $\mathbf{r}_t^{\prime}$ is divided into cubes of $8\times8\times8$ voxels and each cube is separately transformed by using a 3D DCT \cite{zigzag,3ddct}. 
Thereafter, the transform coefficients are quantized using a quantization matrix.

The quantized transform coefficients are entropy coded and transmitted together with auxiliary information such as motion field $\mathbf{M}_t$, frame type, etc.
 Specifically, the DC coefficients are coded using the Differential Pulse Code Modulation (DPCM) method \cite{jpeg}.
\begin{figure*}[t]
	\vspace{-0.5cm}
	\begin{center}
		\includegraphics[width=\linewidth]{ 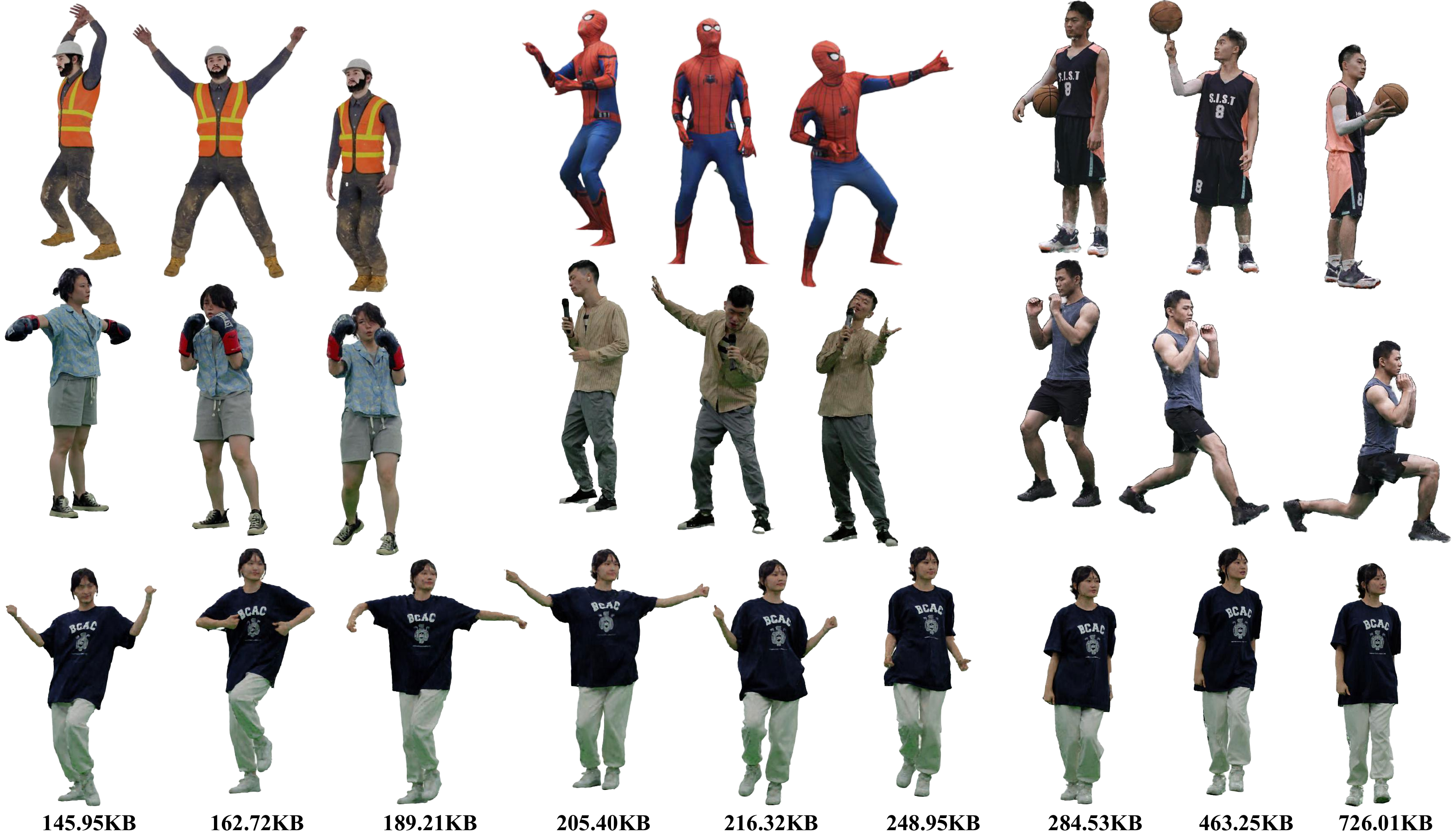}
	\end{center}
	\vspace{-0.6cm}
	\caption{The rendered appearance results of our ReRF method on inward 360\degree~ long sequences with large motions. The last row shows that we can enable variable bitrate.}
	\label{fig:gallery}
	\vspace{-5mm}
\end{figure*}

The AC coefficients coding involves arranging the quantized DCT coefficients in a ``3D zigzag"  order \cite{zigzag}, employing a run-length encoding (RLE) algorithm to group similar frequencies together, inserting length coding zeros.
Finally, we use Huffman coding to further compress the DPCM-coded DC coefficients and the RLE-coded AC coefficients. An advantage of our compression method is the ability to achieve variable bitrates via adjusting the quantization parameters, thus enabling dynamic adaptive streaming of ReRF according to the available bandwidth.

\subsection{Network Streaming ReRF Player}

We also implement a companion ReRF player for online streaming dynamic radiance fields of long sequences, with broad control functions.  When the bitstream is received, the I-feature grid $\mathbf{\widetilde{f}}_1$  is first reconstructed by performing inverse quantization and inverse transform on the quantized transform coefficients.

After the I-feature grid is reconstructed, the subsequently received P-feature grid will then be reconstructed. Specifically, the initial reconstructed residual grid $\mathbf{\hat{r}}_t^{\prime}$ is generated by inverse quantization and inverse transform of the quantized transform coefficients. Then $\mathbf{\hat{r}}_t^{\prime}$ is back-projected to the origin space  $\mathbf{\hat{r}}_t=\mathbf{\hat{r}}_t^{\prime} \cdot \mathbf{V}^T$. Additionally, given the decoded motion field $\mathbf{M}_t$ and the previously reconstructed feature grid $\mathbf{\widetilde{f}}_{t-1}$, we can obtain the predicted feature grid $\mathbf{\overline{f}}_t$ by deformation.
Finally, $\mathbf{\overline{f}}_t$ as well as  $\mathbf{\hat{r}}_t$ are added to produce the final  reconstructed feature grid $\mathbf{\widetilde{f}}_t$. $\mathbf{\widetilde{f}}_t$ is  output  to the renderer to  generate photo-realistic FVV of dynamic scenes.

Benefiting from the design of the GOF  structure, our ReRF player allows fast seeking to a new position to play during playback. Because encountering a new GOF in a compressed bitstream means that the decoder can decode a compressed feature grid without reconstructing any previous feature grid. With ReRF player, for the first time, users can pause, play, fast forward/backward, and seek dynamic radiance fields just like viewing a 2D video, bringing an unprecedented high-quality free-viewpoint viewing experience.

\section{Experimental Results}

In this section, we evaluate our ReRF on a variety of challenging scenarios.
Our captured dynamic datasets contain around 74 views at the resolution of 1920$\times$1080 at 25 fps. 
We use the PyTorch Framework to train the proposed network on a single NVIDIA GeForce RTX3090. We also implement a companion ReRF player for online streaming dynamic fields of long sequences.
To verify the effectiveness of the proposed ReRF player, we use a PC with  Intel(R) Core(TM) i9-11900 CPU@2.5 GHz and NVIDIA GeForce RTX3090 GPU as the test platform.
In the experiments, the length of each GOF is set to 20.
As demonstrated in Fig.~\ref{fig:gallery} and Fig.~4 in the supplementary, we can generate high-quality appearance results in both inward 360\degree and forward-facing scenes with long sequences and large, challenging motions.
Our method can flexibly adjust storage by scaling the quantization factor shown in the third row of  Fig.~\ref{fig:gallery}. Please refer to the supplementary video for more video results.

\begin{figure*}[t]
\vspace{-3mm}
\begin{center}
    \includegraphics[width=\linewidth]{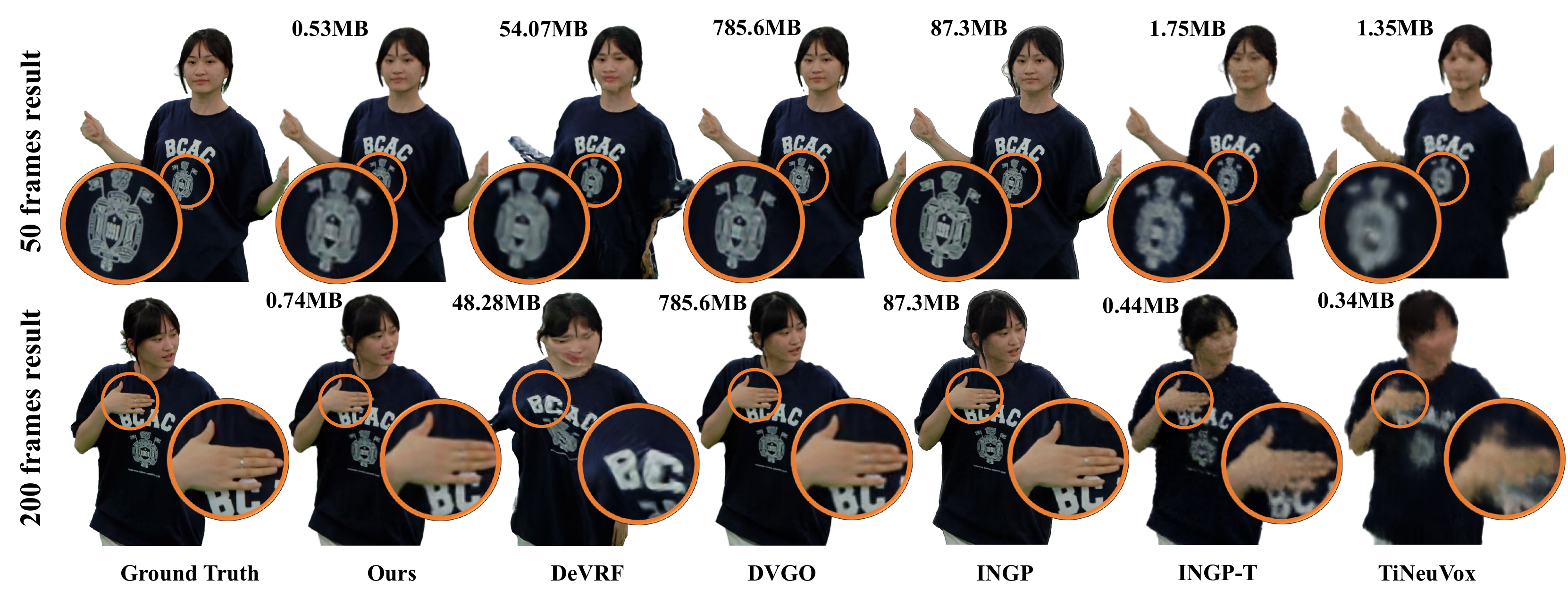}
\end{center}
\vspace{-0.7cm}
\caption{Qualitative comparison against dynamic scene reconstruction methods and per frame static reconstruction methods.
}
\label{fig:comparison}

\end{figure*}

\begin{table*}[t]

\begin{center}
\small
\centering\setlength{\tabcolsep}{6pt}
\renewcommand{\arraystretch}{1.1}

\setlength{\tabcolsep}{1.5mm}{\begin{tabular}{l | ccccc | ccccc} 
\hline
       &  \multicolumn{5}{c|}{50 frames} & \multicolumn{5}{c}{200 frames} \\ \hline
Method & Size(MB)$\downarrow$ & PSNR$\uparrow$ & SSIM$\uparrow$& MAE$\downarrow$ & LPIPS $\downarrow$  & Size(MB)$\downarrow$  & PSNR$\uparrow$ & SSIM$\uparrow$& MAE$\downarrow$ & LPIPS $\downarrow$  \\ 
\hline
DeVRF\cite{liu2022devrf}        & 54.07      & 26.03&0.9508&0.0142&0.0587   & 48.28 & 20.63&0.9192&0.0275&0.0978\\

DVGO\cite{sun2021direct}            & 785.6  & 37.88&0.9922&0.0021&0.0199  & 785.6 & 37.80&0.9920&0.0020&0.0192 \\

INGP\cite{muller2022instant}       & 87.30    & 38.75& 0.9936 & 0.0014 & 0.0192   & 87.30 & 38.86 & 0.9943 & 0.0015 & 0.0189 \\
INGP-T       & 1.746    & 31.72&0.9668&0.0064  &0.0488  & 0.436 & 30.40&0.9683&0.0059&0.0464 \\
TiNeuVox\cite{fang2022fast}        & 1.348    & 27.79&0.9515&0.0097&0.0671  & 0.337 & 25.84&0.9422&0.0131&0.0836 \\
\hline
Ours             & 0.650    & 37.03&0.9902&0.0023&0.0232  & 0.645 & 37.02&0.9902&0.0023&0.0244\\

\hline 
\bottomrule
\end{tabular}
}
\end{center}
\vspace{-5mm}
\caption{Qualitative comparison against dynamic scene reconstruction methods and per frame static reconstruction methods. We calculate the storage averaged among the frames and PSNR averaged among the frames and views. Compared to origin DVGO, our model size is three order smaller and preserves the visual quality.
}
\label{tab:comparison}
\vspace{-4mm}
\end{table*}

\subsection{Comparison}

\begin{figure}[t]
\vspace{-0.2cm}
	\begin{center}
		\includegraphics[width=0.98\linewidth]{ 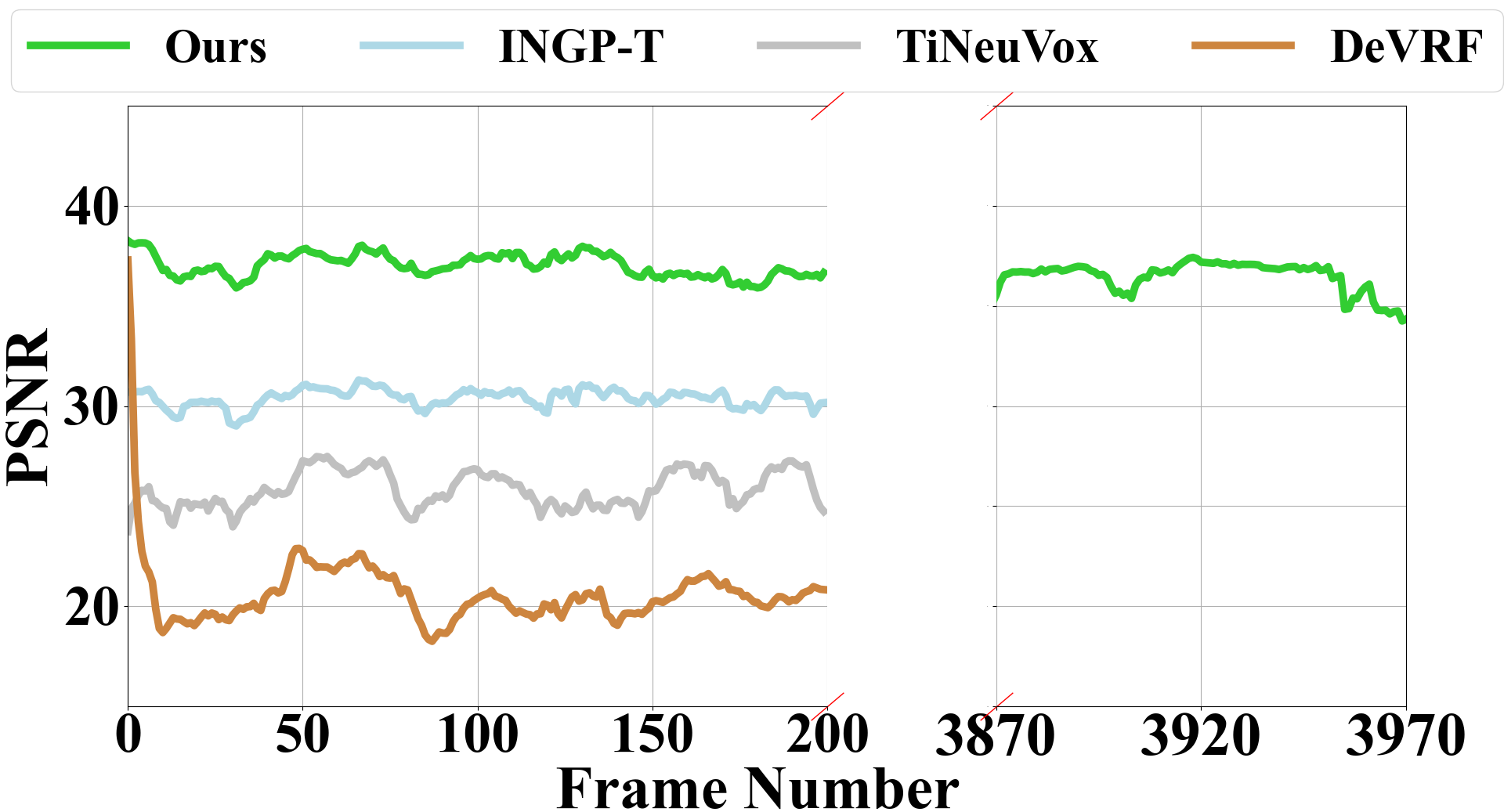}
	\end{center}
	\vspace{-0.5cm}
	\caption{Quantitative comparison on the number of frames. We show that the performance of our method does not decrease as the number of frames increases.}
	\label{fig:comp3}
	\vspace{-7mm}
\end{figure}

\textbf{Dynamic Scene Comparison.} We provide the experimental results to demonstrate the effectiveness of our proposed ReRF method. We compare with other state-of-the-art methods for dynamic scenes including DeVRF\cite{liu2022devrf},  DVGO\cite{sun2021direct}, INGP\cite{muller2022instant}, INGP-T, and TiNeuVox\cite{fang2022fast} both qualitatively and quantitatively. 
 INGP-T is a modified time-conditioned NGP version. It takes normalized 4D input $[x,y,z,t]$ as hash table input.
In Fig. \ref{fig:comparison}, we report the visual quality  results of different methods when compared with our ReRF compression method on both short and long sequences.
 Specifically, our approach can achieve photo-realistic free-viewpoint rendering comparable to per-frame reconstruction DVGO and INGP, but with much less storage overload. 
Compared to dynamic reconstruction methods (DeVRF, INGP-T, TiNeuVox), we achieve the most vivid rendering result in terms of photo-realism and sharpness, which, in addition, without losing performance in long sequences. 
DeVRF learns an explicit deformation field from the live frame to the first frame. When the motion is large, especially in long sequences, it is difficult to warp directly from the first frame. INGP-T and TiNeuVox suffer from severe blurring effects as the frame count increases. Note that no matter how the number of frames increases (even to 4000 frames), our method always maintains high photo-realism and sharpness as shown in Fig. \ref{fig:comp3}.

For quantitative comparison, we adopt the peak signal-to-noise ratio (\textbf{PSNR}), structural similarity index (\textbf{SSIM}) as metrics to evaluate our rendering accuracy. 
We choose 70 captured views as training set and the other 4 views as testing set. In Tab.\ref{tab:comparison}, we show that we can effectively use the small storage to achieve high-quality results. In long sequences with large motions, our method outperforms other dynamic methods in terms of appearance.

Also, note that our method can achieve fast training (about 10 mins per frame) and fast rendering (20fps), significantly faster than NeRF and many previous methods.

\subsection{Evaluation}

\begin{figure}[t]
	\vspace{-0.5cm}
	\begin{center}
		\includegraphics[width=0.98\linewidth]{ 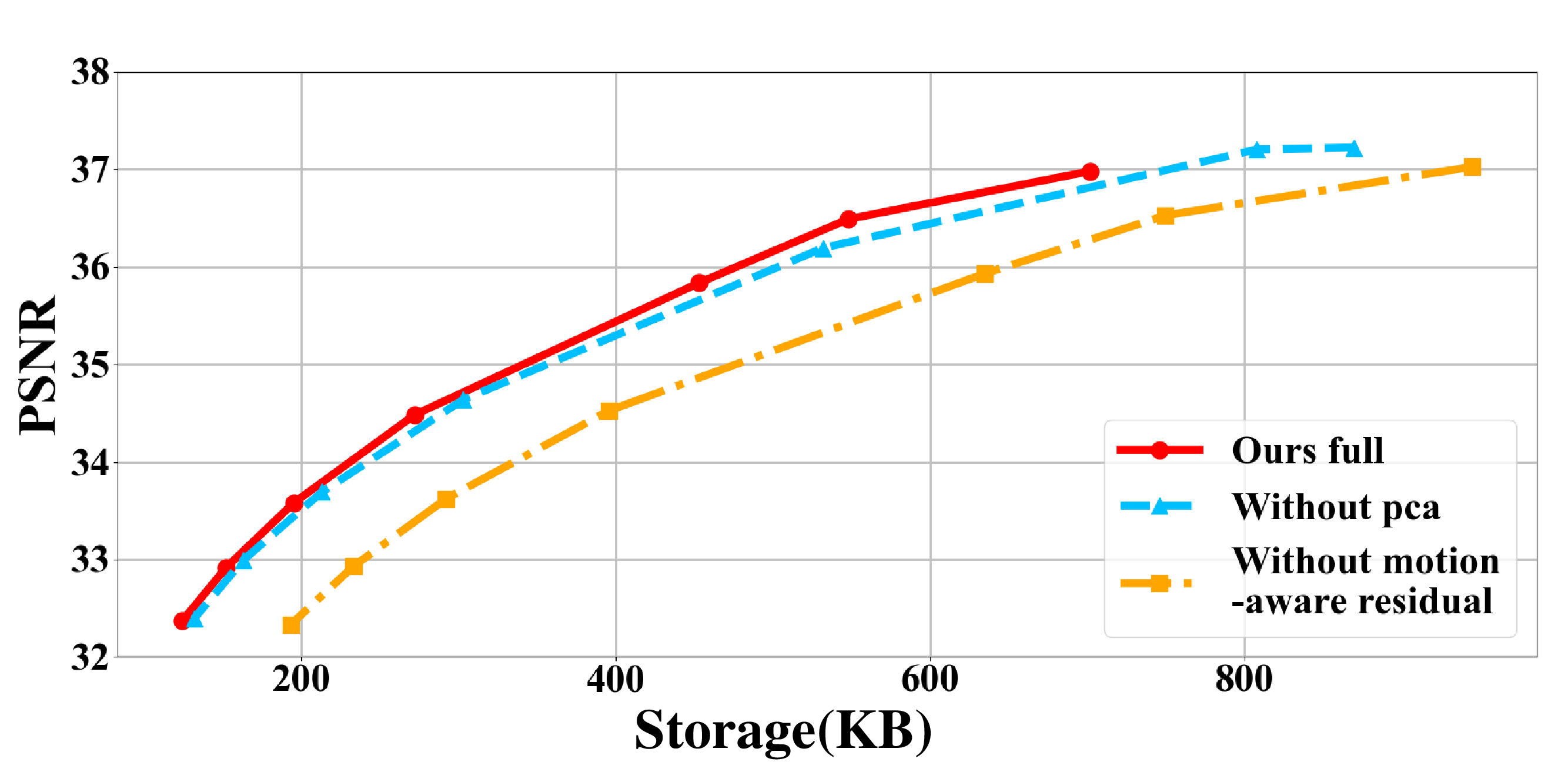}
	\end{center}
	\vspace{-0.7cm}
	\caption{\textbf{Rate distortion curve.} This figure shows the rate distortion of our different components. Our complete architecture is the most compact and is able to dynamically scale the bitrate to different storage requirements.}
	\label{fig:eval1}
	\vspace{-3mm}
\end{figure}

\begin{figure}[t]
	\begin{center}
	    \includegraphics[width=0.98\linewidth]{ 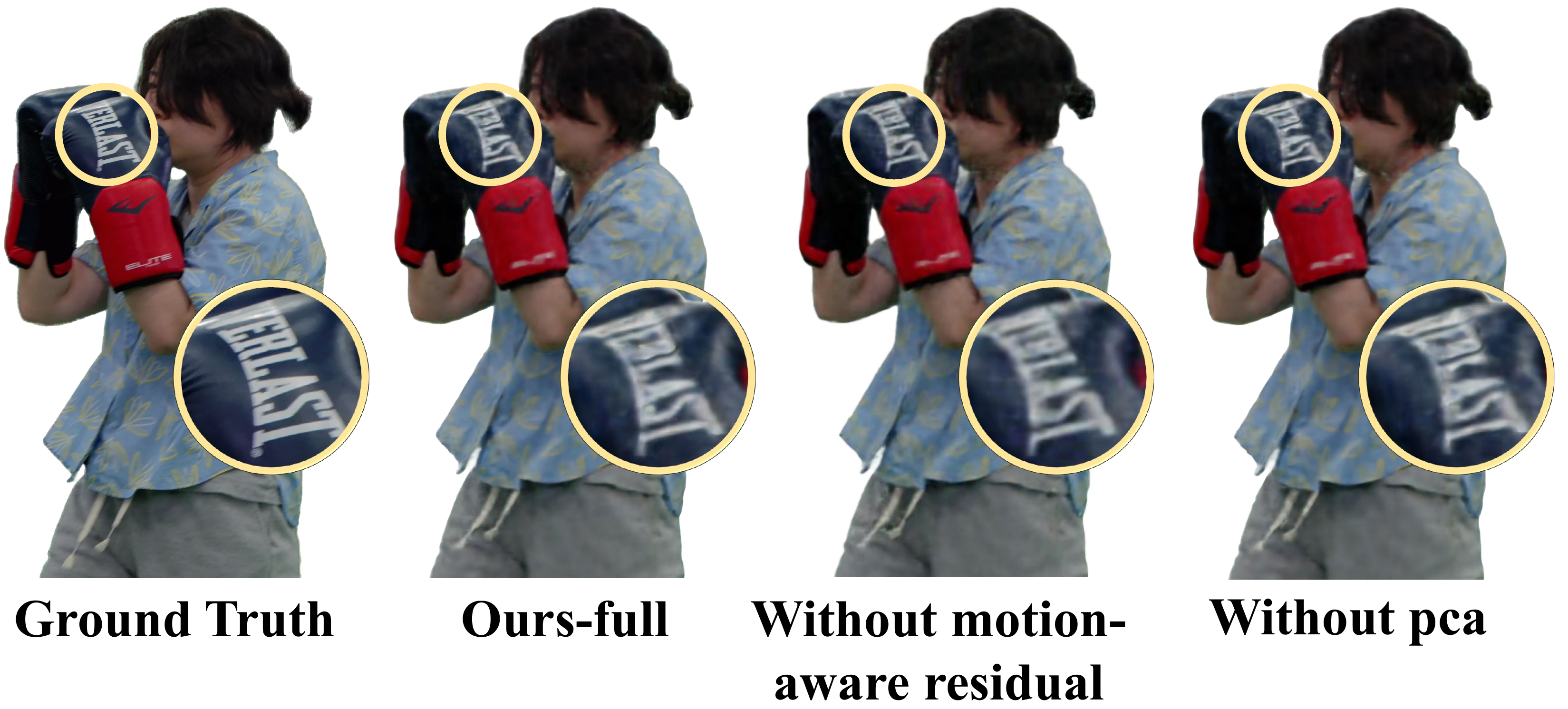}
	\end{center}
	\vspace{-0.5cm}
	\caption{\textbf{Qualitative evaluation of different variations in our method.}}
	\label{fig:eval2}
	\vspace{-5mm}
\end{figure}

\textbf{Ablation Study.} We analyze the motion-aware residual module and our PCA module.
For without motion-aware residual, we train each frame independently and directly encode the residual of 2 frames.
Fig. \ref{fig:eval1} highlights that our motion-aware can significantly improve compactness. Also, our PCA module can improve even further.
In Fig. \ref{fig:eval2}, we show the result under the limit of 700KB storage. In contrast, our complete model generates photorealistic results with minimal noise caused by compression.

\textbf{Analysis of storage.} We show the storage of each component in our high-quality version in Tab. \ref{storage}. We report the average bitrate of our compressed residual feature, voxel motion field, PCA back-project matrix $V^T$  and others including masks to indicate the empty space and header file information. Note that, our total average model size is 793KB which is three orders of magnitude more compact. %

\textbf{Analysis of runtime.}  As shown in the runtime breakdown analysis on Tab. \ref{breakdown}, our ReRF player supports real-time decoding and rendering of on-demand ReRF streams. The average time to decode and render one frame is about 47.03ms and 44.62ms, respectively. 
In addition, the decoding time and rendering time are close to each other, which is more friendly to parallel processing. The total processing time of the player, achieved by decoding and rendering in parallel, is about 50ms. Users can experience free-view videos at high frame rates in an immersive manner, just as smoothly as viewing 2D videos on YouTube.

\begin{table}[t]
\vspace{-0.3cm}
	\centering
	\begin{tabular}{l|c|c|c|c}
		\hline
		Components & Residual & Motion & PCA & others\\ \hline
		Size (KB) & 755.31 &31.80& 0.68& 4.86\\  \hline
		 \multicolumn{2}{l}{Origin Size}&  \multicolumn{3}{c}{786MB}\\\hline	
        \bottomrule
	\end{tabular}
	\vspace{-3mm}
	\caption{Quantitative evaluation on the storage of different Components. We show that our proposed method is 1000 times smaller than the original model size without compression.}
	\vspace{-3mm}
	\label{storage}
\end{table}
\begin{table}[t]
	\centering
	\begin{tabular}{l|c|c}
		\hline
		Stage & Action & Avg Time\\
		\hline
		\multirow{4}{*}{Decoding} 
		&  entropy decoding & $\sim$  26.01 ms \\ 
		&  inverse quantization & $\sim$  0.08 ms \\  
            & 3D IDCT & $\sim$  1.32 ms \\ 
		&  others & $\sim$  19.62 ms \\
            \hline
		Rendering &   - & $\sim$   44.62 ms \\ \hline	
            \bottomrule
	\end{tabular}
	\vspace{-3mm}
	\caption{Breakdown of processing per-frame time in each stage of ReRF player. The result is averaged over a whole sequence.}
	\vspace{-5mm}
	\label{breakdown}
\end{table}

\section{Discussion}

\textbf{Limitation.} As the first trial to enable streamable radiance field modeling and rendering for long sequences with rich experiences, our approach has some limitations.
First, compared to storage, our averaged per-frame training time needs to be improved. We will try some training acceleration techniques from \cite{muller2022instant,li2022streaming}. 
Second, although we have reached 20 fps, speeding up our rendering for more fluent interaction is the direction we need to explore.
Moreover, we need a multiview capture system to provide dynamic sequences, which is expensive and hard to construct. 

\textbf{Conclusion.} We have presented a novel Residual Radiance Field (ReRF) technique for compactly modeling long-duration dynamic scenes.
Our novel motion/residual grids in ReRF are compression-friendly to model the spatial-temporal feature space of dynamic scenes in a sequential manner.
Our ReRF-based codec scheme achieves three orders of magnitude compression improvement, while our ReRF player further enables online dynamic radiance fields streaming and free-viewing.
Our experimental results demonstrate the effectiveness of ReRF for highly compact and effective dynamic scene modeling.
With the unique streamable ability for long-duration dynamic scenes, we believe that our approach serves as a critical step for neural scene modeling, with various potential immersive applications in VR/AR. 

\section{Acknowledgements}
This work was supported by Shanghai YangFan Program (21YF1429500), Shanghai Local college capacity building program (22010502800), NSFC programs (61976138, 61977047), STCSM (2015F0203-000-06), and SHMEC (2019-01-07-00-01-E00003). We also acknowledge support from Shanghai Frontiers Science Center of Human-centered Artificial Intelligence (ShangHAI). We thank Penghao Wang, Zhipeng He and Zhirui Zhang for their assistance in  the experiments and figures.
{\small
\bibliographystyle{ieee_fullname}
\bibliography{egbib}
}

\clearpage

\section{Appendix}

\appendix

\section{Training details for Neural Residual Field.}
\begin{figure}[t]
	\vspace{-0.1cm}
	\begin{center}
		\includegraphics[width=1.05\linewidth]{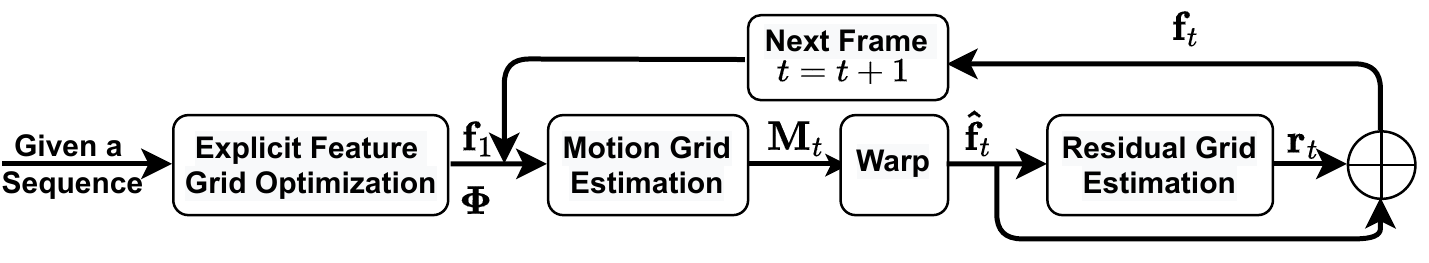}
	\end{center}
	\vspace{-0.2cm}
	\caption{Overview for the training of ReRF.}
	\label{fig:flow}
	\vspace{-2mm}
\end{figure}

Here we provide a detailed illustration of the training scheme to generate ReRF from RGB video inputs, as shown in Fig.~\ref{fig:flow}.
Specifically, we construct an explicit feature grid for the first frame. Then, sequentially given the former feature grid $\mathbf{f}_{t-1}$ and the RGB images at the current timestamp, we optimize our motion grid $\mathbf{M}_t$ and residual grid $\mathbf{r}_t$ for the current frame to generate our compact neural representation.

{\let\thefootnote\relax\footnote{\
		${ }^{\dagger}$ The corresponding authors are Minye Wu
		(minye.wu@kuleuven.be) and Lan Xu (xulan1@shanghaitech.edu.cn).
}}\par

\subsection{Feature Grid Optimization at the First Frame.}
Given a long-duration multi-view sequence, we first learn an explicit feature grid and a global MLP $\Phi$ from the first frame.
Similar to DVGO~\cite{sun2021direct}, we use an explicit density grid $\mathbf{V}_{\sigma}$ and a color feature grid $\mathbf{V}_{c}$ to represent the first frame. 
To render a view, we will cast rays through the pixels and sample points along rays.
For the sampled point $\mathbf{p}$, we will query the scene property (density and color feature) efficiently through trilinear interpolation from the grids:

$$
\text { Tri-Interp }(\mathbf{p}=[x, y, z], \mathbf{V}_{\sigma}):\left(\mathbb{R}^3, \mathbb{R}^{ N_x \times N_y \times N_z}\right) \rightarrow \mathbb{R},
$$
\begin{equation}
	\text { Tri-Interp }(\mathbf{p}=[x, y, z], \mathbf{V}_{c}):\left(\mathbb{R}^3, \mathbb{R}^{C \times N_x \times N_y \times N_z}\right) \rightarrow \mathbb{R}^C,
\end{equation}
where $C$ is the number of color feature dimension,$~ N_x,~ N_y$ and $N_z$ are the voxel resolutions of $\mathbf{V}_{\sigma}$ and $\mathbf{V}_{c}$. We choose $C=12$ in our experiments. 
Following DVGO, we adopt the softplus and post-activation to obtain the density property of the sample points and apply the global MLP $\Phi$ to the color feature for the view-dependent rendering. This shallow MLP contains two hidden layers and each layer is 128 channels.
During the training, we progressive upscale our density grid $\mathbf{V}_{\sigma}$ and color feature grid $\mathbf{V}_{c}$. The initial number of voxels is  $125\times125\times 125$. After reaching the training step 1000, 2000 and 4000, the final resolution will be upscaled to $150\times 150 \times 150$, $200\times 200 \times 200$ and $250\times 250 \times 250$, respestively. 

During training this explicit feature grid, we employ the photometric MSE loss and apply total variation loss on $\mathbf{V}_{\sigma}$:

$$
\mathcal{L}_{render_{explicit}}= \sum_{\mathbf{l} \in \mathbb{L}}\|\mathbf{c}(\mathbf{l})-\mathbf{\hat c}(\mathbf{l})\|,
$$
$$
\mathcal{L}_{TV_{explicit}}=\frac{1}{|\mathbf{V_{\sigma}}|} \sum_{\substack{\mathbf{v} \in \mathbf{V_{\sigma}} }} \sqrt{\Delta_x^2\mathbf{v}+\Delta_y^2\mathbf{v}+\Delta_z^2\mathbf{v}},
$$
\begin{equation}
	\mathcal{L}_{explicit}=\mathcal{L}_{render_{explicit} }+\lambda_{\text {TV}} \mathcal{L}_{TV_{explicit} },
\end{equation}

where $\lambda_{\text {TV}}=0.000016$; $\mathbb{L}$ is the set of training pixel rays; $\mathbf{c}(\mathbf{l})$ and $\mathbf{\hat c}(\mathbf{l})$ are the ground truth color and predicted color of a ray $\mathbf{l}$ respectively. $\Delta_{x,y,z}^2\mathbf{v}$ denotes the squared difference between the density value in the voxel. The total variation loss is only activated during the training iteration 1000 to 12000.
We utilize the Adam optimizer for training 5000 iterations in the coarse stage and 16000 iterations in the fine stage with a batch size of 10192 rays. The learning rate for $\mathbf{V}_{\sigma}$, $\mathbf{V}_{c}$ and global MLP is 0.1, 0.11 and 0.002, respectively.

\subsection{Motion Grid Optimization.}
Here we provide details to generate our compact low-resolution motion grid $\mathbf{M}_t$, which represents the position offset from the current frame to the previous so as to exploit feature similarities.
We propose to generate $\mathbf{M}_t$ from a densely estimated motion field $\mathbf{D}_t$, which is a grid with a shape of $3\times N_x \times N_y \times N_z$ and contains the warping information from the frame $t$ to frame $t-1$.

For our dense motion field estimation, we first sample point $\mathbf{p}_{t}$ along the pixel ray of frame $t$. Then the sampled point $\mathbf{p}_{t}$ will query the 3D motion $\Delta \mathbf{p}_{t\rightarrow t-1}=\mathbf{D}_t(\mathbf{p}_{t})$ through trilinear interpolation:

\begin{equation}
	\text { Tri-Interp }(\mathbf{p}_{t}=[x, y, z], \mathbf{D}_t):\left(\mathbb{R}^3, \mathbb{R}^{ 3 \times N_x \times N_y \times N_z}\right) \rightarrow \mathbb{R}^3.
\end{equation}

After finding the corresponding point $\mathbf{p}_{t-1}=\mathbf{p}_t+\Delta \mathbf{p}_{t\rightarrow t-1}$, we can get the feature from the previous feature grid $\mathbf{f}_{t-1}$ for $\mathbf{p}_t$. Then, the global MLP $\Phi$ will decode color feature to RGB space. Finally, the pixel color can be calculated through volume rendering.

During this estimation, we also progressively upscale the deformation field $\mathbf{D}_t$ from $(125\times125\times 125)\rightarrow (150\times150\times 150)\rightarrow  (200\times200\times 200)\rightarrow (250\times250\times 250)$ after reaching the training step 1000, 2000 and 4000, respectively. We adopt the following photometric MSE loss and total variation loss to estimate $\mathbf{D}_t$:

$$
\mathcal{L}_{render_{deform}}=\sum_{\mathbf{l} \in \mathbb{L}}\|\mathbf{c}(\mathbf{l})-\mathbf{\hat c}(\mathbf{l})\|,
$$
$$
\mathcal{L}_{TV_{deform}}=\frac{1}{|\mathbf{D}_t|} \sum_{\substack{\mathbf{v} \in \mathbf{D}_t }} \sqrt{\Delta_x^2(\mathbf{v},d)+\Delta_y^2(\mathbf{v},d)+\Delta_z^2(\mathbf{v},d)},
$$
\begin{equation}
	\mathcal{L}_{deform}=\mathcal{L}_{render_{deform} }+\lambda_{\text {TV}} \mathcal{L}_{TV_{deform} },
\end{equation}

where the total variation loss enforces the smoothness of the dense motion field, and $\lambda_{\text {TV}}$ is set to be 1. We use the Adam optimizer for training 3000 iterations in the coarse stage and 16000 iterations in the fine stage, with a batch size of 10192 rays and a learning rate of $10^{-4}$.

Then, we generate the smooth and compact motion grid $\mathbf{M}_t$ from $\mathbf{D}_t$ through a motion pooling strategy as described in the main manuscript.

\subsection{Residual Grid Optimization.}
Here we provide implementation details to generate the sparse residual grid $\mathbf{r}_{t}$ of the current frame, which is used to compensate for warping errors and newly observed regions. 
Specifically, we first warp the previous frame feature $\mathbf{f}_{t-1}$ using the compact motion field $\mathbf{M}_t$ to generate a base feature grid $\mathbf{\hat f}_{t}$.
Then, during the optimization, we shoot rays from the image pixels and sample points $\mathbf{p}_t$ along it. The base feature grid and residual grid are both queried through trilinear interpolation to obtain $ \mathbf{\hat f}_{t}(\mathbf{p}_t)$ and $ \mathbf{r}_{t}(\mathbf{p}_t)$:

$$
\text { Tri-Interp }(\mathbf{p}_t=[x, y, z], \mathbf{\hat f}_{t}):\left(\mathbb{R}^3, \mathbb{R}^{C \times N_x \times N_y \times N_z}\right) \rightarrow \mathbb{R}^C,
$$
\begin{equation}
	\text { Tri-Interp }(\mathbf{p}_t=[x, y, z], \mathbf{r}_{t}):\left(\mathbb{R}^3, \mathbb{R}^{C \times N_x \times N_y \times N_z}\right) \rightarrow \mathbb{R}^C.
\end{equation}
Note that $C=13$, since we union the density and color feature in our feature grid $\mathbf{f}$ representation. 
We obtain the final scene property of the current frame through the summation: $\mathbf{f}_t(\mathbf{p}_t) = \mathbf{\hat f}_{t}(\mathbf{p}_t) +\mathbf{r}_t(\mathbf{p}_t)$. Finally, the global MLP $\Phi$ will decode it into radiance fields with volume rendering to calculate pixel color.

We employ the same progressive training scheme for the residual grid, starting from  $(125\times125\times 125)\rightarrow (150\times150\times 150)\rightarrow  (200\times200\times 200)\rightarrow (250\times250\times 250)$ after reaching the training step 1000, 2000 and 4000, respectively. 
Besides the photo-metric MSE loss and  total variation loss on density residual, we utilize an additional L1 loss to encourage the residual sparsity:

$$
\mathcal{L}_{render_{residual}}=\sum_{\mathbf{l} \in \mathbb{L}}\|\mathbf{c}(\mathbf{l})-\mathbf{\hat c}(\mathbf{l})\|,
$$
$$
\mathcal{L}_{TV_{residual}}=\frac{1}{|\mathbf{f_{t}^d}|} \sum_{\substack{\mathbf{v} \in \mathbf{f_{t}^d} }} \sqrt{\Delta_x^2\mathbf{v}+\Delta_y^2\mathbf{v}+\Delta_z^2\mathbf{v}},
$$
$$
\mathcal{L}_{1{residual}}=\frac{1}{|\mathbf{r}_t|} \sum_{\substack{\mathbf{v} \in \mathbf{r}_t }} (|\mathbf{v}_x|+|\mathbf{v}_y|+|\mathbf{v}_z|),
$$
\begin{equation}
	\mathcal{L}_{residual}=\mathcal{L}_{render_{residual} }+\lambda_{\text {TV}} \mathcal{L}_{TV_{residual} }+\lambda_{\text {residual }} \cdot \mathcal{L}_{\text {1residual}}, 
\end{equation}

where $\mathbf{f_{t}^d}$ represents the density of feature grid $\mathbf{f_{t}}$; $\lambda_{\text {TV}}=0.000016$ and $\lambda_{\text {residual }}=0.01$.

Similar to our first frame explicit grid optimization, the total variation loss of density residual is only activated during the training iteration 1000 to 12000.
We adopt the Adam optimizer for training 5000 iterations in the coarse stage and 16000 iterations in the fine stage with a batch size of 10192 rays. The learning rate for the density of residual grid $\mathbf{r_{t}}$, and the color feature of residual grid $\mathbf{r_{t}}$ is 0.1, 0.11, respectively. Note that the base feature grid $\mathbf{\hat f}_{t}$ is fixed and we initialize the  residual grid with zero value during our residual grid optimization.

\section{ReRF Codec and Streamble Application}\label{sec:4}

In the past decades, a number of image and video compression standards like JPEG\cite{jpeg}, JPEG2000\cite{jpeg2000}, H.264/AVC\cite{h264} and H.265/HEVC\cite{h265} have been proposed and widely used in many practical applications.  Most of these video compression methods follow a hybrid coding structure, where motion compensation and residual coding are adopted to reduce spatial and temporal redundancy. Recent work has also attempted to employ neural networks for video compression and has shown considerable performance\cite{codec1,codec2,codec3,codec4,codec5,codec6}. 
Inspired by these compression methods, we propose a ReRF-based codec and a companion FVV player for online streaming of long-duration dynamic scenes, while still guaranteeing an immersive exploration experience over existing networks. Fig. \ref{fig:encoder} demonstrates the overall pipeline of our framework. 

\begin{figure}[t]
	\vspace{-0.1cm}
	\begin{center}
		\includegraphics[width=1.0\linewidth]{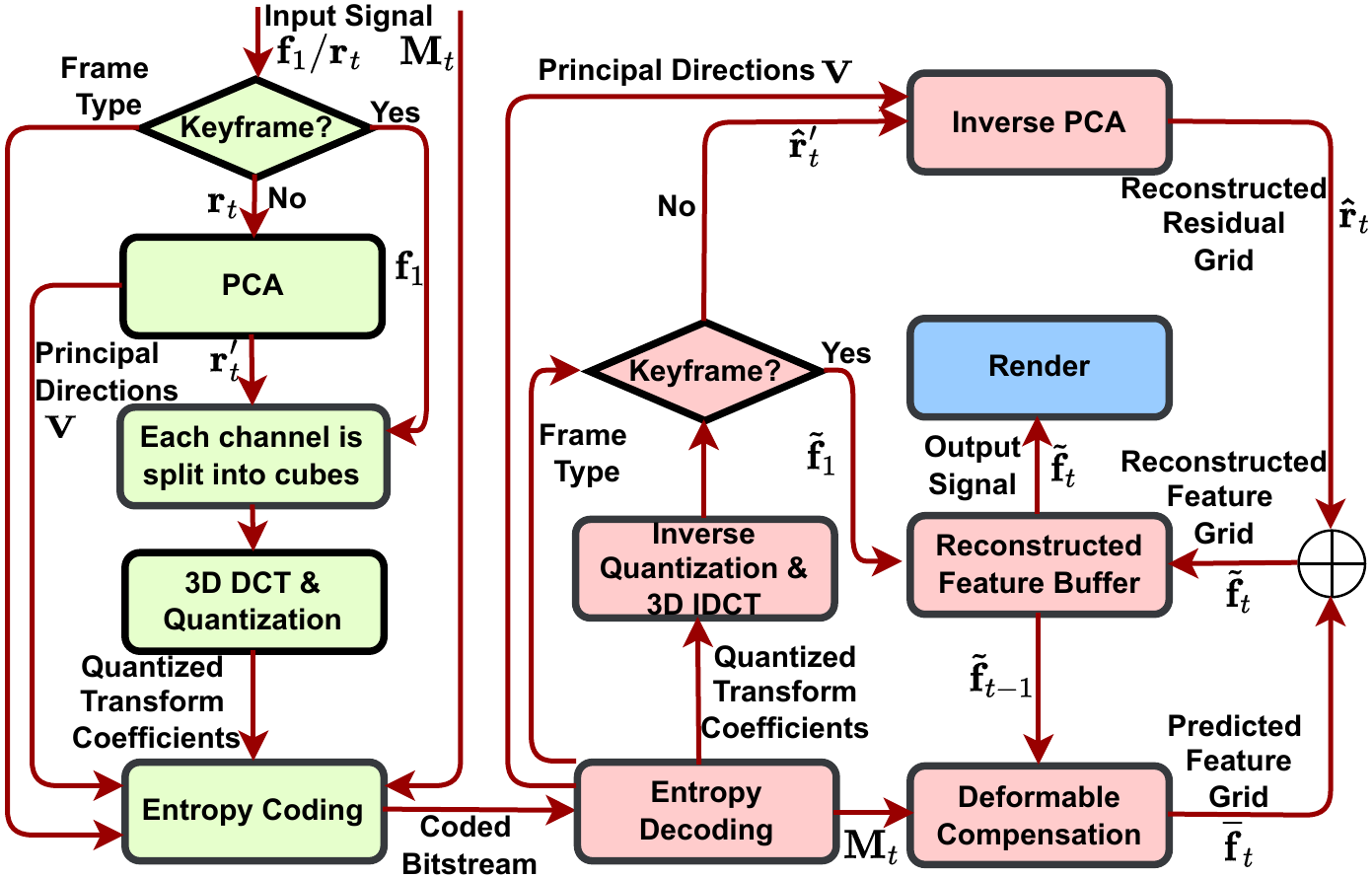}
	\end{center}
	\vspace{-0.2cm}
	\caption{Overview of our proposed ReRF-based codec and player (the modeling elements of the encoder and decoder are shaded in light green and pink, respectively). The encoder compresses the input signal to produce a bitstream by using PCA, 3D-DCT, quantization, and entropy coding. The decoder receives the compressed bitstream, decodes each of the syntax elements, and reverses the coding process. Additionally, given the decoded motion field $\mathbf{M}_t$ and the previously reconstructed feature grid $\mathbf{\widetilde{f}}_{t-1}$, we can obtain the predicted feature grid $\mathbf{\overline{f}}_t$ by deformation.}
	\label{fig:encoder}
	\vspace{-2mm}
\end{figure}

\subsection{Feature-level Residual Compression.}
\label{sec:4.1}
Both motion and residual grids are amenable to compression, especially for long-duration dynamic scenes.
To make ReRF practical for users, we design a ReRF-based codec that follows the traditional keyframe-based strategy.
We first divide the feature grid sequence into several continuous groups of feature grids (GOF), which is a collection of successive grids as shown in Fig \ref{GOF}.
GOFs are comprised of two frame types, the I-feature grid (keyframe) and the P-feature grid. 
Each GOF begins with an I-feature grid which is coded independently of all other feature grids and contains most of the vital information for the following sequence of the P-feature grid.
The p-feature grid contains a deformable compensated residual grid relative to the previous feature grid.
Let $\{\mathbf{f}_1,\mathbf{r}_2,\cdots,\mathbf{r}_{t-1},\mathbf{r}_t,\cdots\}$ denote a GOF, 
where $\mathbf{f}_1$ is the original feature grid and $\mathbf{r}_t$ is the residual grid at the current time step. 
Our goal is to generate high quality reconstructed feature grid $ \mathbf{\widetilde{f}}_t$ at any given bitrate.

\textbf{PCA.}
We first reshape $\mathbf{f}_1$ and 
$\mathbf{r}_t$ into $\mathbf{f}_1(m,n)$ and $\mathbf{r}_t(m,n)$, a $m\times n$ feature matrix, where $m$ and $n$ are the number of non-empty feature voxels and feature channels, respectively. %
Then, we perform linear Principal Component Analysis (PCA) \cite{pca} on $\mathbf{r}_t(m,n)$ to obtain the named-tuple $(\mathbf{U},\mathbf{S},\mathbf{V})$ which is the nearly optimal approximation of a singular value decomposition of $\mathbf{r}_t(m,n)$: 
\begin{equation}
	\label{pca1}
	\mathbf{r}_t(m,n)=\mathbf{U} \cdot \operatorname{diag}(\mathbf{S}) \cdot \mathbf{V}^T,
\end{equation}
where $\mathbf{V}$ is the $n\times q$ matrix, representing the principal directions, $\mathbf{S}$ is the $q$-vector, $\mathbf{U}$ is the $m\times q$ matrix. 
Finally, we project the $\mathbf{r}_t$ to principal directions as follows:
\begin{equation}
	\label{pca}
	\mathbf{r}_t^{\prime}=\mathbf{r}_t\cdot \mathbf{V}.
\end{equation}

\textbf{3D DCT.}
Each channel of grid $\mathbf{f}_1$ and $\mathbf{r}_t^{\prime}$ is divided into cubes of $8\times8\times8$ voxels and each cube is separately transformed by using a 3D DCT \cite{zigzag,3ddct}.
Let residual voxels for each cube are denoted by $r(i,j,k)$, and the DCT coefficients $R(u,v,w)$ can be calculated as:

\begin{equation}
	\label{eqn:3d-dct}
	\begin{aligned}
		R(u, v, w) =& C_u C_v C_w\sum_{i=0}^{N-1}\sum_{j=0}^{N-1}\sum_{k=0}^{N-1}r(i,j,k)\cos{[\dfrac{(2i+1)\pi}{2N}u]} \\
		&\cos{[\dfrac{(2j+1)\pi}{2N}v]}\cos{[\dfrac{(2k+1)\pi}{2N}w]}\\
		&\text{where}  \   \   C_u,C_v,C_w=\left\{\begin{aligned}
			&\sqrt{\dfrac{1}{N}} \ \ i,j,k=0 \\
			&\sqrt{\dfrac{2}{N}} \ \ \text{otherwise.} \\
		\end{aligned}
		\right.
	\end{aligned}
\end{equation}

The transformed value at the coordinate origin $R(0,0,0)$ is the DC coefficient which is the most important value of the transformed coefficients. The amplitude of the DC coefficient is larger and contains more energy. While the rest coefficients are the AC coefficients, they contain little energy in the whole 3D-DCT, and most of their energy is concentrated on the major axis of the cube.  

\begin{figure}[t]
	\vspace{-0.1cm}
	\begin{center}
		\includegraphics[width=1.0\linewidth]{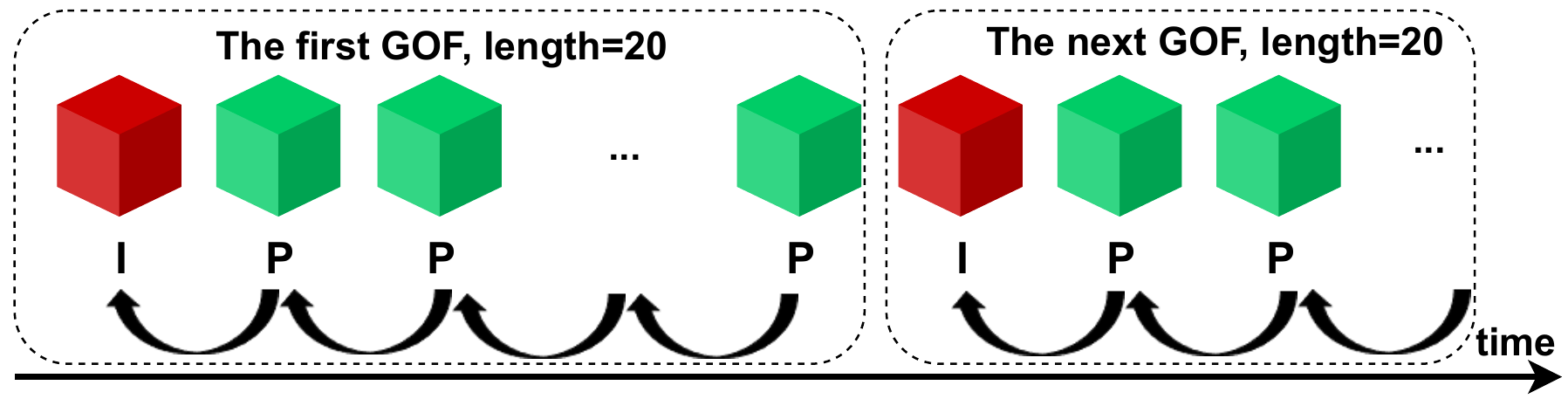}
	\end{center}
	\vspace{-0.2cm}
	\caption{ GOF structure. }
	\label{GOF}
	\vspace{-2mm}
\end{figure}

\textbf{Quantization.}
Thereafter, the transform coefficients are quantized using a quantization matrix. The quantization matrix for a coefficient cube should have an entry for each coefficient. The values in the quantization matrix depend on if the corresponding coefficients are significant and also on the underlying quality factor being adopted. We perform scalar quantization on the 3D DCT coefficients. Each quantized transform coefficient is given by 
\begin{equation}
	\label{eqn:quantization}
	\hat{R}(u,v,w) = round(\dfrac{R(u,v,w)}{S_q\times Q(u,v,w)}).
\end{equation}
where $S_q$ is a scaling factor and $Q(u,v,w) $ is a quantization matrix entry.  In this work, we construct a quantization matrix based on psycho-visual experiments. The values of the quantization matrix are provided in quant.npy in the supplementary material.

\textbf{Entropy Coding.}
The quantized transform coefficients are entropy coded and transmitted together with auxiliary information such as motion field $\mathbf{M}_t$, frame type, etc.
Entropy coding involves arranging the quantized DCT coefficients in a ``3D zigzag"  order \cite{zigzag}, employing a run-length encoding (RLE) algorithm to group similar frequencies together, inserting length coding zeros, and then using Huffman coding \cite{jpeg,huffman} for the remainder. 
The DC  coefficients are coded separately from AC  ones \cite{jpeg}.
Specifically, the DC coefficients are coded using the Differential Pulse Code Modulation (DPCM) method \cite{jpeg}: except for the first DC coefficient, we encode the difference between the current DC coefficient and the previous DC coefficient. 
The AC coefficients are coded using the RLC method. To make it most likely hit a long run of zeros, a ``3D zig-zag" scan is adopted. Finally, we use Huffman coding to further compress the DPCM-coded DC coefficients and the RLE-coded AC coefficients.

The experimental results show that our ReRF-based codec achieves three orders of magnitudes compression rate compared to per-frame-based neural representations ~\cite{sun2021direct}.
Another advantage of our compression method is the ability to achieve variable bitrates via adjusting the scaling factor $S_q$ during quantization, thus enabling dynamic adaptive streaming of ReRF according to the available bandwidth.

\subsection{Network Streaming ReRF Player}

We also implement a companion ReRF player for online streaming dynamic radiance fields of long sequences, with broad control functions.  Our ReRF player supports downloading the coded bitstream from streaming media servers.  When the bitstream is received, the I-feature grid $\mathbf{\widetilde{f}}_1$  is first reconstructed by performing inverse quantization and inverse transform on the quantized transform coefficients.

After the I-feature grid is reconstructed, the subsequently received P-feature grid will then be reconstructed. Specifically, the initial reconstructed residual grid $\mathbf{\hat{r}}_t^{\prime}$ is generated by inverse quantization and inverse transform of the quantized transform coefficients. Then $\mathbf{\hat{r}}_t^{\prime}$ is back-projected to the origin space by
\begin{equation}
	\label{eqn:ipca}
	\mathbf{\hat{r}}_t=\mathbf{\hat{r}}_t^{\prime} \cdot \mathbf{V}^T.
\end{equation}

Additionally, given the decoded motion field $\mathbf{M}_t$ and the previously reconstructed feature grid $\mathbf{\widetilde{f}}_{t-1}$, we can obtain the predicted feature grid $\mathbf{\overline{f}}_t$ by deformation. Let $\mathbf{p}$ denote the index of our explicit grids. Then, the predicted feature grid is formulated as:
\begin{equation}
	\label{eqn:deformation}
	\mathbf{\overline{f}}_t(\mathbf{p}) = \mathbf{\widetilde{f}}_{t-1}(\mathbf{p}+\mathbf{M}_t(\mathbf{p})).
\end{equation}

The predicted feature grid $\mathbf{\overline{f}}_t$ as well as the initial reconstructed residual grid $\mathbf{\hat{r}}_t$ are added to produce the final  reconstructed feature grid $\mathbf{\widetilde{f}}_t$, as follows:
\begin{equation}
	\label{eqn:rec}
	\mathbf{\widetilde{f}}_t = \mathbf{\overline{f}}_t+\mathbf{\hat{r}}_t.
\end{equation}

Finally, the  reconstructed feature grid $\mathbf{\widetilde{f}}_t$ is  output  to the renderer to  generate photo-realistic FVV of dynamic scenes. 
As our ReRF player can efficiently reconstruct and render dynamic scenes,  users are free to choose their views as if they were in the target scene.

Benefiting from the design of the GOF  structure, our ReRF player allows fast seeking to a new position to play during playback. The reason is that the coded bitstream consists of successive GOFs. The first frame in a GOF is an I-feature grid (keyframe) which contains an independently coded feature. Encountering a new GOF in a compressed bitstream means that the decoder can decode a compressed feature grid without reconstructing any previous feature grid. With ReRF player, for the first time, users can pause, play, fast forward/backward, and seek dynamic radiance fields just like viewing a 2D video, bringing an unprecedented high-quality free-viewpoint viewing experience.

Note that the I-feature grid (keyframe) is different from the explicit feature grid for the first frame. We only use explicit feature grid to representation in the first frame training. All other sequential frame features are trained using residual grid $\mathbf{r}_t$ and can be generated to feature grid $\mathbf{f}_t$. GOF structure is used to enable fast seeking. It will choose key frame every GOF size. For I-feature grid, the full feature grid $\mathbf{f}_t$ is encoded (generated from residual grid) . For P-feature grid, the residual grid $\mathbf{r}_t$ is encoded.

\begin{figure}[t]
	\begin{center}
		\includegraphics[width=0.95\linewidth]{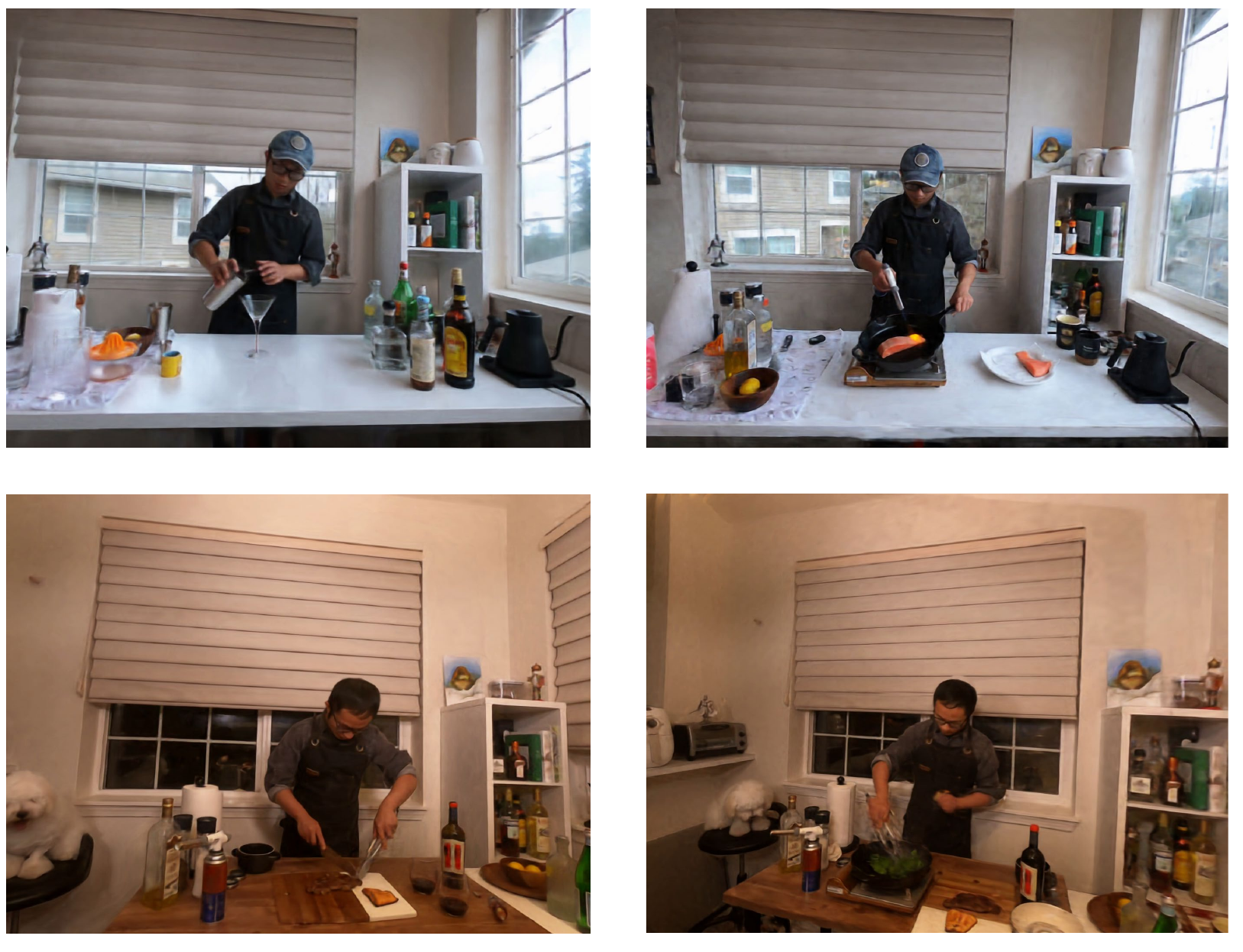}
	\end{center}
	\vspace{-0.7cm}
	\caption{Our rendering results for forward facing scenes in neural 3D dataset.}
	\label{fig:gallery2}

\end{figure}

\begin{table}[t]
	\vspace{-0.3cm}
	\begin{center}
		\small
		\centering\setlength{\tabcolsep}{6pt}
		\renewcommand{\arraystretch}{1.1}
		\setlength{\tabcolsep}{1.5mm}{
			\begin{tabular}{l | c | cc | cc} 
				\hline
				&           & \multicolumn{2}{c|}{Synthetic-NeRF} & \multicolumn{2}{c}{TanksTemples}    \\ \hline
				Method & Size$\downarrow$ & PSNR$\uparrow$ & SSIM$\uparrow$              & PSNR$\uparrow$ & SSIM$\uparrow$  \\ 
				\hline
				SRN  \cite{sitzmann2019scene}      & -      & $22.26$ & $0.846$   & $24.10$ & $0.847$ \\
				NeRF \cite{mildenhall2020nerf}      & $5.0$    & $31.01$ & $0.947$   & $25.78$ & $0.864$ \\
				NSVF  \cite{liu2020neural}         & -      & $31.75$ & $0.953$   & $28.48$ & $0.901$ \\
				SNeRG  \cite{hedman2021snerg}       & $1771.5$  & $30.38$ & $0.950$   & -     & -     \\
				POctrees \cite{yu2021plenoctrees} & $1976.3$ & $31.71$ & -       & $27.99$ & $0.917$ \\
				Plenoxels \cite{fridovich2022plenoxels}     & 778.1  & $31.71$ & -       & $27.43$ & $0.906$ \\
				DVGO   \cite{sun2021direct}         & $612.1$   & $31.95$ & $0.975$   & $28.41$ & $0.911$ \\
				TRF-CP  \cite{chen2022tensorf}     & $3.9$    & $31.56$ & $0.949$   & $27.59$ & $0.897$ \\
				TRF-VM  \cite{chen2022tensorf}     & $71.8$    & $33.14$ & $0.963$   & $28.56$ & $0.920$ \\
				INGP\cite{muller2022instant} & $63.3$    & $33.18$ & -       & -     & -     \\
				CC-CP  \cite{tang2022compressible}     & $4.4$    & $30.55$ & $0.935$   & $27.01$ & $0.878$ \\
				CC-HY  \cite{tang2022compressible}      & $88.0$    & $32.37$ & $0.955$   & $28.08$ & $0.913$ \\
				\hline
				Ours (high)                            & 3.21    & 31.81 & 0.955   & 28.30 & 0.910 \\
				Ours (low)                            & 0.98    & 30.14 & 0.941   & 27.16 & 0.893 \\
				\hline
				\bottomrule
			\end{tabular}
		}
	\end{center}
	\vspace{-5mm}
	\caption{Comparison with recent methods on static scenes. We compare our method with previous and concurrent novel view
		synthesis methods on two datasets. All scores of the baseline methods are
		directly taken from their papers whenever available. Our method use minimal storage while maintaining a high PSNR.
	}
	\label{tab:static}
	\vspace{-5mm}
\end{table}

\begin{figure}[t]
	\begin{center}
		\includegraphics[width=1.0\linewidth]{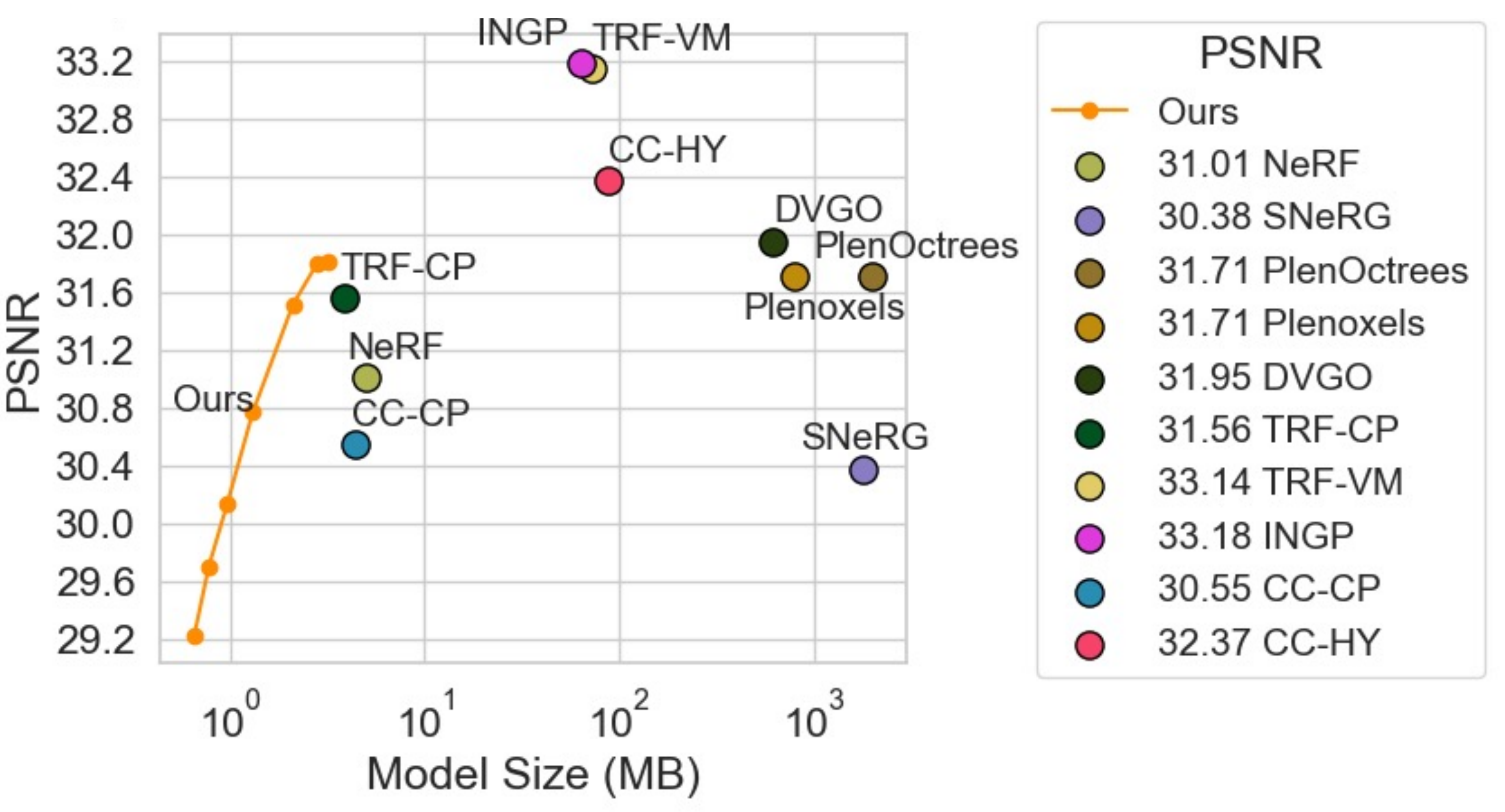}
	\end{center}
	\vspace{-0.3cm}
	\caption{ Quantitative results on Synthetic NeRF Dataset. }
	\label{comp1}
\end{figure}
\section{Experiments}

\subsection{Dataset Details}
Our captured dynamic datasets contain around 74 views at the resolution of 1920×1080 at 25fps. The cameras are the cylindrical distribution looking at the center. Most sequences are more than 1000 frames, The longest sequence contains 4000 frames. We use five real-world captured data and two synthetic data for experiments.

\subsection{Additional Experimental Results}

\textbf{Static Scene Comparison.} 
To demonstrate our  I-feature grid (keyframe) compression performance, we also compare it with the existing static scene novel view synthesis approaches on the Synthetic NeRF dataset \cite{mildenhall2020nerf} and TanksTemples dataset\cite{knapitsch2017tanks} in Tab. \ref{tab:static}. 
Compared to the original DVGO, orders of magnitudes smaller bitrates are achieved, without significantly sacrificing quality.
We choose 2 different quantization factors to show our high-quality compression and low-quality compression results. 
Note that our high quality version has achieved the most compact modeling with essentially the same rendering quality as the original DVGO and also outperforms vanilla NeRF and many other methods.
Our low-quality version also demonstrates that we can use a much more compact storage ($<$1MB) to reach a high PSNR($>$30).

\begin{figure}[t]
	\vspace{0.3cm}
	\begin{center}
		\includegraphics[width=1.0\linewidth]{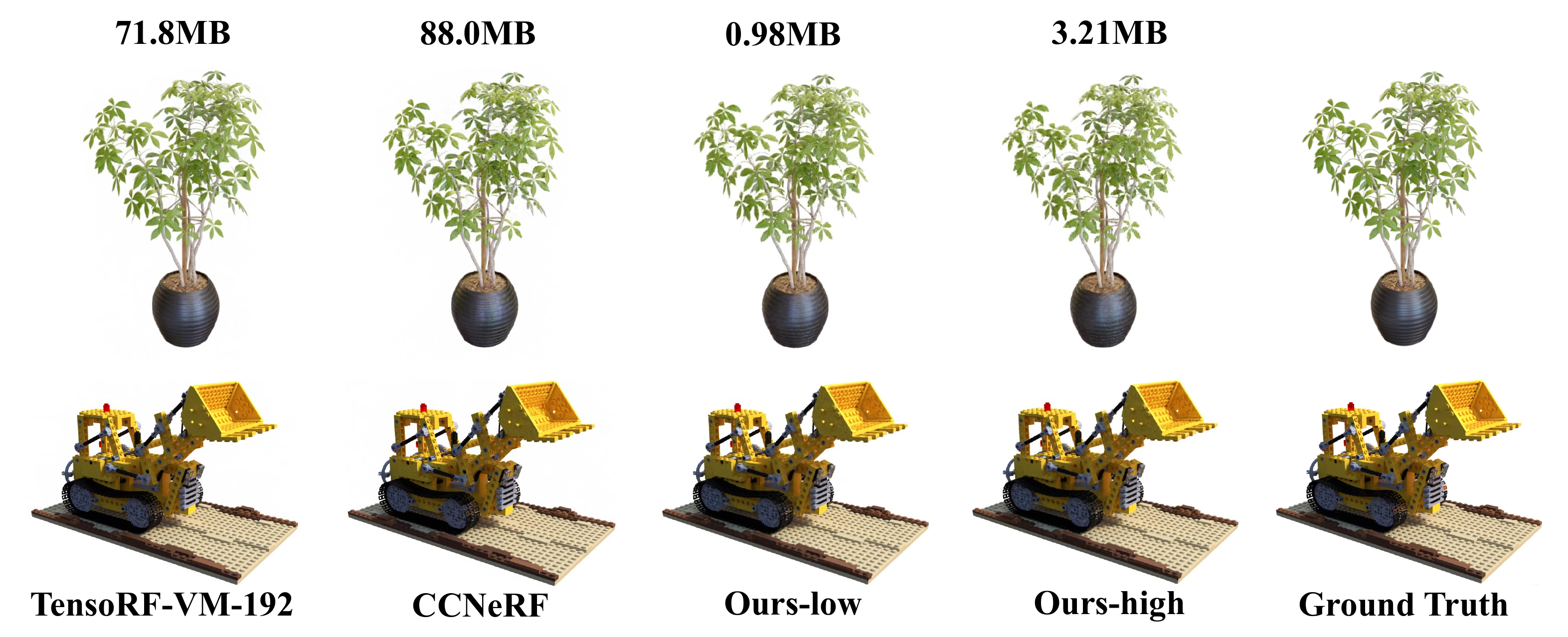}
	\end{center}
	\vspace{-0.6cm}
	\caption{ Qualitative comparisons with several recent works. }
	\label{comp2}
	\vspace{-3mm}
\end{figure}

\begin{figure}[t]
	\begin{center}
		\includegraphics[width=1.0\linewidth]{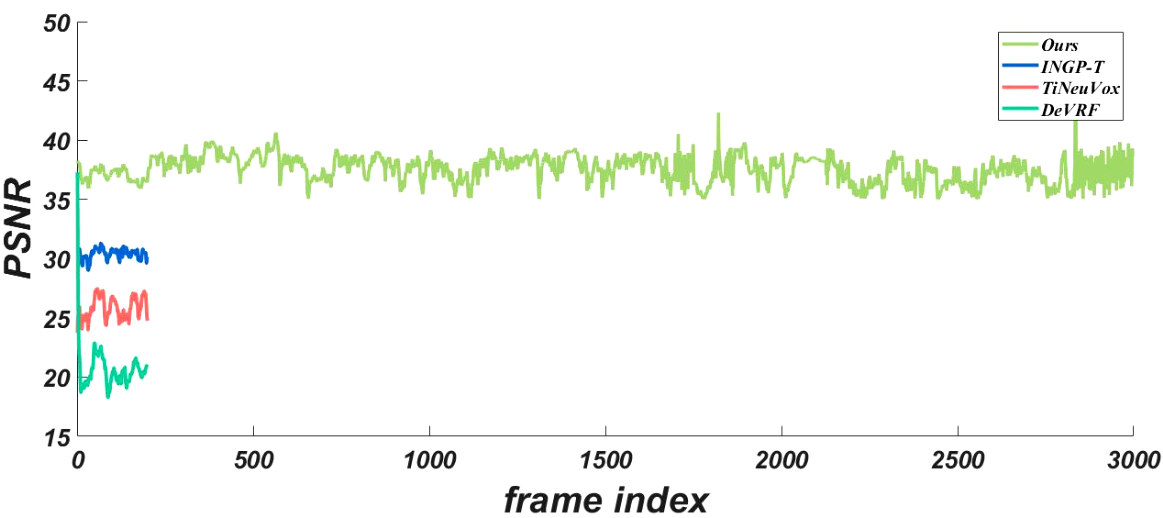}
	\end{center}
	\vspace{-0.6cm}
	\caption{ Performance on the long sequences (3000 frames). }
	\label{comp3}
	\vspace{-6mm}
\end{figure}

Fig. \ref{comp1} and \ref{comp2} further show that our method is the most compact and maintains a high render quality on static scenes in comparison with previous and concurrent methods. Seven different quantization scaling factors are adopted to achieve variable bitrates. Our methods also simultaneously achieve fast reconstruction and rendering.

\end{document}